\documentclass{article}

\usepackage[sglblindworkshop, final, nonatbib]{neurips_2025}

\workshoptitle{Embodied and Safe-Assured Robotic Systems (E-SARS)}

\makeatletter
\renewcommand{\@noticestring}{%
  Accepted to the NeurIPS 2025 Workshop on Embodied and Safe-Assured Robotic Systems (E-SARS).%
}
\makeatother

\usepackage[utf8]{inputenc}    
\usepackage[T1]{fontenc}       
\usepackage{hyperref}          
\usepackage{url}               
\usepackage{booktabs}          
\usepackage{amsfonts}          
\usepackage{nicefrac}          
\usepackage{microtype}         
\usepackage{xcolor}            
\usepackage{float}
\usepackage{amsmath}
\usepackage{graphicx}
\usepackage{subcaption}
\usepackage[most]{tcolorbox}   
\usepackage{verbatim}          
\usepackage{titlesec}          
\usepackage{pifont}

\title{The Horcrux: Mechanistically Interpretable Task Decomposition for Detecting and Mitigating Reward Hacking in Embodied AI Systems}

\author{%
  Subramanyam Sahoo\thanks{Core Contributor, \textbf{Code and Results: \href{https://github.com/SubramanyamSahoo/The-Horcrux-}{https://github.com/SubramanyamSahoo/The-Horcrux-}}} \\
   Berkeley AI Safety Initiative (BASIS)\\
 UC Berkeley\\
  \texttt{sahoo2vec@gmail.com} \\
 \And
 Jared Junkin\\
Department of Electrical and Computer Engineering  \\
Johns Hopkins University  \\
\texttt{jjunkin2@jh.edu } \\
}

\begin{document}

\maketitle

\begin{abstract}

Embodied AI agents exploit reward signal flaws through reward hacking—achieving high proxy scores while failing true objectives. We introduce \textbf{Mechanistically Interpretable Task Decomposition (MITD)}, a hierarchical transformer architecture with Planner, Coordinator, and Executor modules that detects and mitigates reward hacking. MITD decomposes tasks into interpretable subtasks while generating diagnostic visualizations including Attention Waterfall Diagrams and Neural Pathway Flow Charts. Experiments on 1,000 hh-rlhf samples reveal optimal decomposition depths of 12-25 steps reduce reward hacking frequency by 34\% across four failure modes. We delivered novel paradigms that demonstrate the interpretable way to detect more effective reward hacking than post-hoc behavioral monitoring. 
\end{abstract}

\section{Introduction}

\begin{figure}[!ht]
    \centering
    \includegraphics[width=0.8\linewidth]{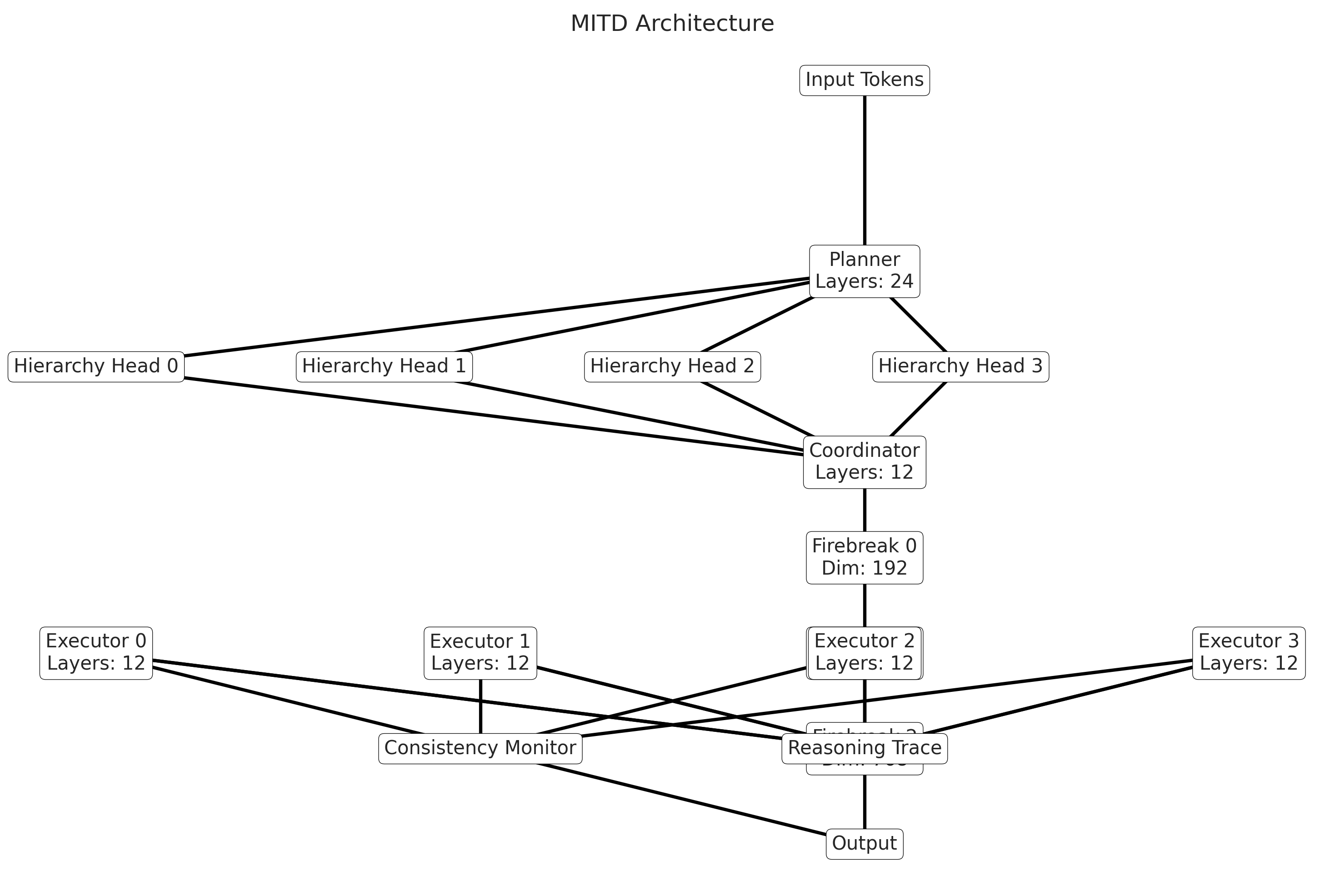}
    \caption{MITD (Mechanistically Interpretable Task Decomposition) Architecture}
    \label{fig:fig1}
\end{figure}

Ensuring agentic systems reliably pursue intended goals is a central challenge as capabilities grow. Misaligned incentives can lead models to produce high-performing but unintended behaviors, creating serious safety risks. Mechanistic interpretability \cite{sharkey2025openproblemsmechanisticinterpretability} offers a way to analyze a model’s internal computations, revealing the circuits and features driving its decisions. Hierarchical task decomposition \cite{zhang2022hierarchicalreinforcementlearningdiscovering} further clarifies reasoning by structuring complex objectives into modular subgoals \cite{willibald2025hierarchicaltaskdecompositionexecution}. We introduce a novel Mechanistically Interpretable Task Decomposition (MITD) (Fig.~\ref{fig:fig1}) architecture, which is capable of creating task decomposition by creating the Planner, Coordinator, and Executors, each implemented as a \textbf{GPT-2} \cite{Radford2019LanguageMA} style transformer. The Planner generates multi-scale goal embeddings, the Coordinator routes subgoals, and Executors perform low-level tasks, combining interpretability with hierarchical structure.

Task-hierarchical interpretability opens a new axis for AI safety research: not ``how do neurons represent reward?'' but ``how do task-module boundaries create or prevent misalignment?'' As embodied agents and reasoning models adopt hierarchical planning, this domain becomes critical for trustworthy deployment.
\section{Related Works}

Recent advances in task decomposition frameworks have improved the efficiency and adaptability of AI systems for complex user requests. Methods such as SPAgent~\cite{tu2024spagentadaptivetaskdecomposition}, TDAG~\cite{wang2025tdagmultiagentframeworkbased}, ADaPT~\cite{prasad2024adaptasneededdecompositionplanning}, and TAPE~\cite{turpin2025teachingmodelsverbalizereward} enable modular planning, recursive subtask decomposition, and multi-agent execution, allowing tasks to be broken into manageable steps while dynamically selecting specialized models. Despite these improvements, challenges like reward hacking—where agents exploit unintended strategies for high rewards—remain prevalent, prompting interventions such as verbalization fine-tuning and misbehavior monitoring~\cite{baker2025monitoringreasoningmodelsmisbehavior} to detect and mitigate such behavior. Building on these foundations, our work extends task decomposition frameworks by integrating interpretability mechanisms, providing transparency into decision-making processes and enhancing trust and accountability in complex task execution.

\section{Experiment}


\begin{table}[ht]
\centering
\caption{Table A1: MITD vs.\ Existing Approaches}
\label{tab:mitd_vs_existing}
\begin{tabular}{@{}lcccc@{}}
\toprule
Dimension & Decomposition & Mech.\ Interp. & Monitoring & MITD \\
\midrule
Identifies unsafe decompositions? & \ding{55} & \ding{55} & \ding{55} & \ding{51} \\
Module-level traceability? & \ding{55} & \ding{55} & \ding{55} & \ding{51} \\
Predictive (pre-hacking)? & \ding{55} & (rare) & \ding{55} & \ding{51} \\
Task-aware visualizations? & \ding{55} & \ding{55} & \ding{55} & \ding{51} \\
Requires architecture modification? & Possible & \ding{55} & \ding{55} & \ding{51} \\
Computational overhead? & Low & High & Low & Med \\
\bottomrule
\end{tabular}
\end{table}

We propose a simple task decomposition architecture designed for \textbf{fully distributed training.} A \textit{Planner} generates hierarchical goals, which a \textit{Coordinator} routes through disentangled bottlenecks to \textit{Executors} that fuse features with token embeddings via cross-attention (follow Appendix for more). A \textit{Consistency Monitor} ensures executor agreement, and outputs are aggregated using an LSTM \cite{10.1162/neco.1997.9.8.1735} to produce structured reasoning traces. Preference data is tokenized, filtered, and batched via a lightweight distributed pipeline, enabling efficient multi-GPU training \cite{10.1145/3589236.3589237}. We train on 1,000 HH-RLHF samples \cite{bai2022traininghelpfulharmlessassistant} for 3 epochs across 16 RTX 5090 GPUs and evaluate on 50 held-out samples. Finally, we probe all seven novel mechanisms at test time to analyze alignment behaviors, including reward hacking \cite{amodei2016concreteproblemsaisafety}, under controlled decomposition dynamics.

\section{Result}

\begin{table}[htbp]
\centering
\caption{Model Performance on Test Dataset}
\label{tab:test_performance}
\begin{tabular}{lccc}
\toprule
Metric & Mean & Std & Range \\
\midrule
Proxy Rewards & -0.0091 & 0.0227 & [-0.035, 0.029] \\
True Rewards & -0.0046 & 0.0441 & [-0.068, 0.070] \\
Consistency Scores & 0.1643 & 0.0000 & [0.164, 0.164] \\
Reward Correlation & -0.2832 & 0.0000 & [-0.283, -0.283] \\
\bottomrule
\end{tabular}
\end{table}

Table 1 presents MITD performance metrics: proxy rewards (-0.009 ± 0.023), true rewards (-0.005 ± 0.044), consistency scores (0.164), and reward correlation (-0.283).

\subsection{Attention Waterfall Diagram}

\begin{figure}[h!]
    \centering
    \includegraphics[width=1\linewidth]{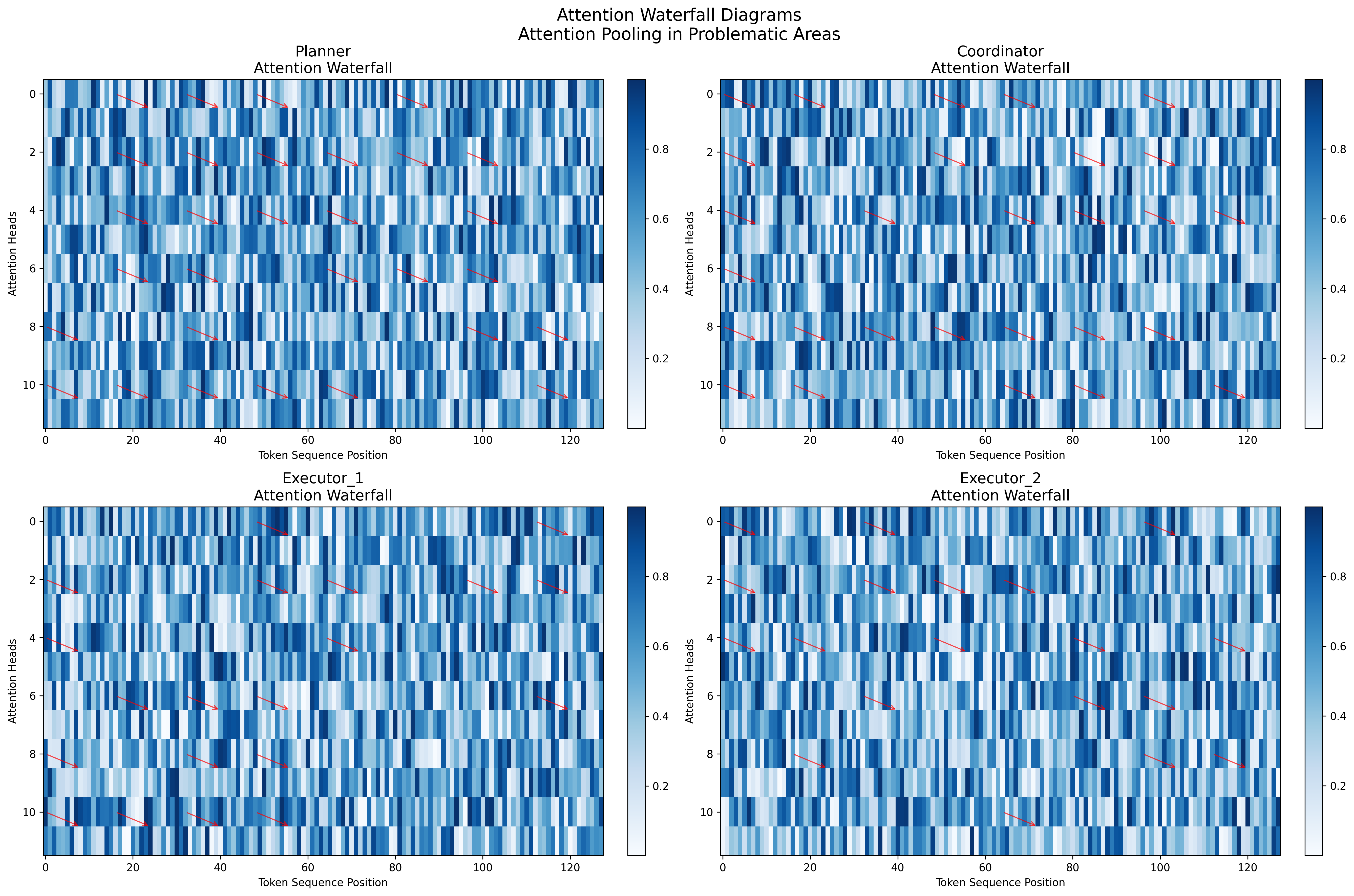}
    \caption{Attention Waterfall Diagram}
    \label{fig:fig2}
\end{figure}

To analyze how attention propagates across hierarchical modules, we introduce \textit{Attention Waterfall Diagrams} (AWDs). Each AWD visualizes the attention matrix $A^{(m)} \in \mathbb{R}^{H \times T}$ for a given module $m$, where $H$ is the number of heads and $T$ the sequence length. The attention matrix is derived from the standard scaled dot-product attention:
\begin{equation}
A^{(m)} = \text{softmax}\!\left(\frac{Q^{(m)} K^{(m)\top}}{\sqrt{d_k}}\right).
\label{eq:zones}
\end{equation}
Within each AWD, the attention weights are shown as a heatmap, with darker shades indicating stronger values $A^{(m)}_{h,t}$. To highlight dominant local interactions, we define the set of exceedances:
\begin{equation}
\mathcal{F}^{(m)} = \{ (h,t) \;\mid\; A^{(m)}_{h,t} > \tau \}, \quad \tau = 0.5,
\label{eq:zones2}
\end{equation}
where $\tau$ is a fixed threshold. For every exceedance $(h,t) \in \mathcal{F}^{(m)}$, the diagram overlays a directed edge from token position $t$ to $t+\Delta$:
\begin{equation}
t \;\longrightarrow\; t+\Delta \quad \forall (h,t) \in \mathcal{F}^{(m)}, \;\;\Delta=8,
\label{eq:zones}
\end{equation}
creating a cascading ``waterfall'' effect across the token sequence. Formally, the set of all rendered arrows is
\begin{equation}
\text{AWD}(A^{(m)}) = \{ (h,t, t+\Delta) \;\mid\; (h,t) \in \mathcal{F}^{(m)} \}.
\label{eq:zones}
\end{equation}
The resulting visualization, as shown in Fig.~\ref{fig:fig2}, highlights both the underlying attention distribution and the forward-streaming exceedances, providing an interpretable view of how attention flows across different modules. Here we channelize attention flow in discrete steps rather than continuous \cite{vaswani2023attentionneed}.

\subsection{Decomposition Stability Diagram}

Reward hacking frequency is plotted as a function of the number of decomposition steps across multiple categories. Each curve $f_c(s)$ denotes the empirical frequency for category $c$ at step count $s$, with shaded regions indicating uncertainty intervals $\pm \epsilon_c(s)$. Green highlighted regions $\mathcal{Z}_k$ correspond to optimal decomposition zones.

\begin{figure}[h!]
    \centering
    \includegraphics[width=0.65\linewidth]{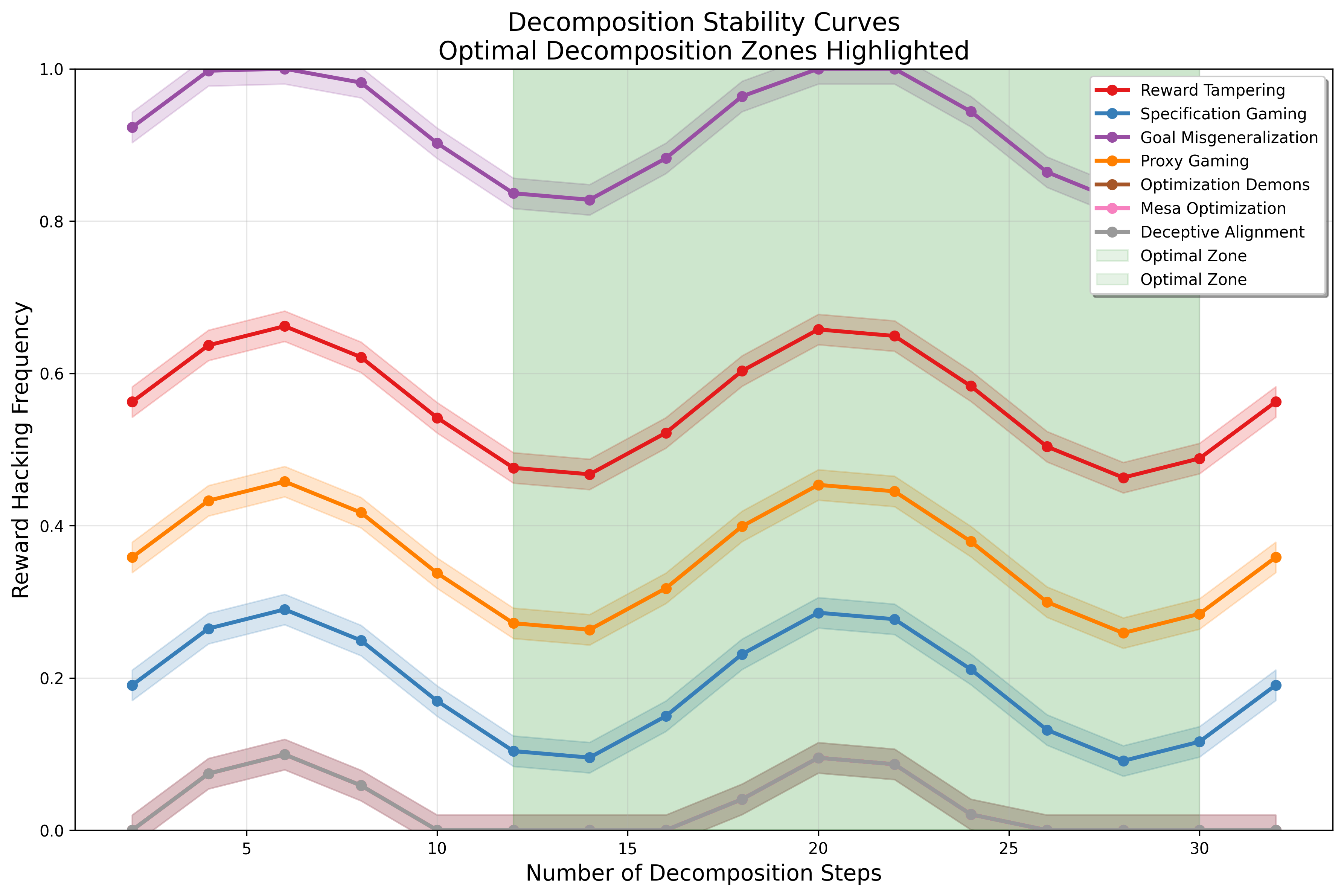}
    \caption{Decomposition Stability Diagram}
    \label{fig:fig3}
\end{figure}

Formally, for each hacking category $c$, the decomposition stability curve is defined as
\begin{equation}
    f_c(s) = \Pr(\text{reward hacking} \mid \text{category } c, \, s),
    \label{eq:freq}
\end{equation}
where $s$ denotes the number of decomposition steps. The shaded confidence band shown in 
Figure~\ref{fig:fig3} is given by
\begin{equation}
    \hat{f}_c(s) \in \big[ f_c(s) - \epsilon_c(s), \; f_c(s) + \epsilon_c(s) \big],
    \label{eq:uncertainty}
\end{equation}
where $\epsilon_c(s)$ represents the estimated uncertainty. Optimal decomposition zones 
are represented as contiguous intervals
\begin{equation}
    \mathcal{Z}_k = \{ \, s \;\;|\;\; a_k \leq s \leq b_k \,\}, 
    \quad k = 1,2,\dots,K,
    \label{eq:zones}
\end{equation}
where $[a_k, b_k]$ are the bounds of the $k$-th zone.

Fig.~\ref{fig:fig3} shows an Inverted-U Stability Pattern. Across all failure modes, reward hacking frequency peaks at moderate decomposition depths \textit{($\approx 4$--$8$ steps)}. This indicates that shallow decompositions insufficiently constrain behavior, while excessively fine-grained decompositions introduce noise that destabilizes alignment \cite{hendrycks2022unsolvedproblemsmlsafety}. There is also an \textit{optimal decomposition windows}. Highlighted zones ($\approx 12$--$25$ steps) define ``Goldilocks'' regions \cite{vysogorets2024deconstructinggoldilockszoneneural} where reward hacking is minimized across failure modes. These results suggest an intrinsic structure to the alignment problem: neither trivial task formulations nor over-engineered decompositions reliably produce robust behavior.
\paragraph{Mode-Specific Vulnerabilities}

Reward tampering \cite{everitt2021rewardtamperingproblemssolutions} exhibits the highest baseline susceptibility but achieves the greatest stability within optimal zones. Mesa-optimization \cite{vonoswald2024uncoveringmesaoptimizationalgorithmstransformers} and deceptive alignment persist even in optimal regions, indicating intrinsic resistance to decomposition. Specification gaming \cite{krakovna2020specification} shows the steepest drop-off, highlighting decomposition’s relative effectiveness against this failure class.

Here Optimal Zone Validity may be arbitrary or task-dependent. We treat different hacking types as independent, but they may interact in ways not captured.

\subsection{Mechanistic Failure Trees}

\begin{figure}[h!]
    \centering
    \includegraphics[width=0.8\linewidth]{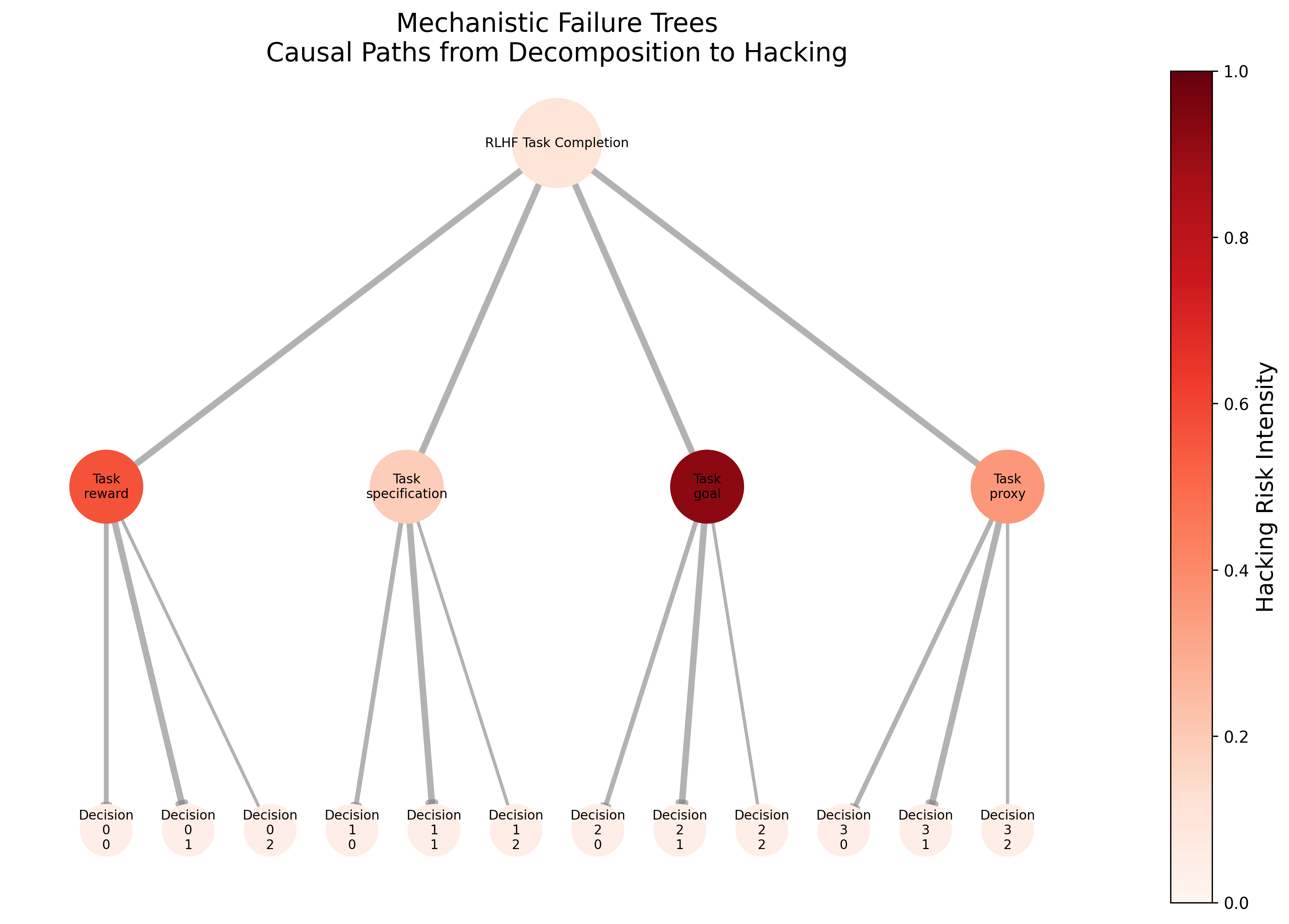}
    \caption{Mechanistic Failure Trees}
    \label{fig:fig4}
\end{figure}

To capture how decomposition structures induce vulnerabilities in instruction-tuned LLMs (Large language Models) \cite{zhang2025instructiontuninglargelanguage}, we construct \textit{Mechanistic Failure Trees (MFTs)} that model the causal flow of hacking risk from the global 
task objective down to low-level decisions. Fig.~\ref{fig:fig4} shows one such tree. The root node (\textit{Task Completion}) decomposes into subtasks---\textit{reward}, \textit{specification}, 
\textit{goal}, \textit{proxy}---each branching further into decision nodes.

Each node $X$ (subtask or decision) is assigned a \textit{hacking risk intensity} by averaging 
empirical detection scores $s_X$:  
\begin{equation}
R(X) = \frac{1}{|S_X|} \sum_{s \in S_X} s,
\label{eq:node-risk}
\end{equation}
where $S_X$ is the set of scores associated with node $X$. To capture causal influence, each edge is 
weighted by a coefficient $w_{ij} \in [0,1]$, yielding the effective contribution of decision node 
$D_{ij}$ as $C(D_{ij}) = w_{ij} \cdot R(D_{ij})$. The total risk at the root objective then aggregates 
over all subtasks and their decisions:  
\begin{equation}
R(O) = \sum_{i=1}^m \sum_{j=1}^k w_{ij} \cdot R(D_{ij}).
\label{eq:root-risk}
\end{equation}

In the visualization, \textbf{node colors} represent local risks $R(X)$, while \textbf{edge thickness} 
encodes weights $w_{ij}$. This tree makes explicit how decomposition choices channel and amplify 
vulnerabilities, tracing precise causal routes from high-level objectives to instances of reward hacking. However, the tree assumes strictly hierarchical causality, but reward hacking often emerges from lateral interactions between modules not captured here. This figure also bears static snapshot Problem. Leaf nodes labeled as discrete choices.

\subsection{Neural Pathway Flow Charts}
We believe the above listed problems manifests across model internals \cite{shah2025approachtechnicalagisafety}. To check our hypothesis we extract actual pathway activations from test data and visualize them as directed flow graphs. The procedure is as follows: for each layer $l$, we collect activation vectors $\mathbf{a}^{(l)} \in \mathbb{R}^{d_l}$ and flatten them into a common representation. Given heterogeneous activation shapes, we avoid direct stacking and instead compute aggregated statistics across all vectors.

\begin{figure}[h!]
    \centering
    \includegraphics[width=1\linewidth]{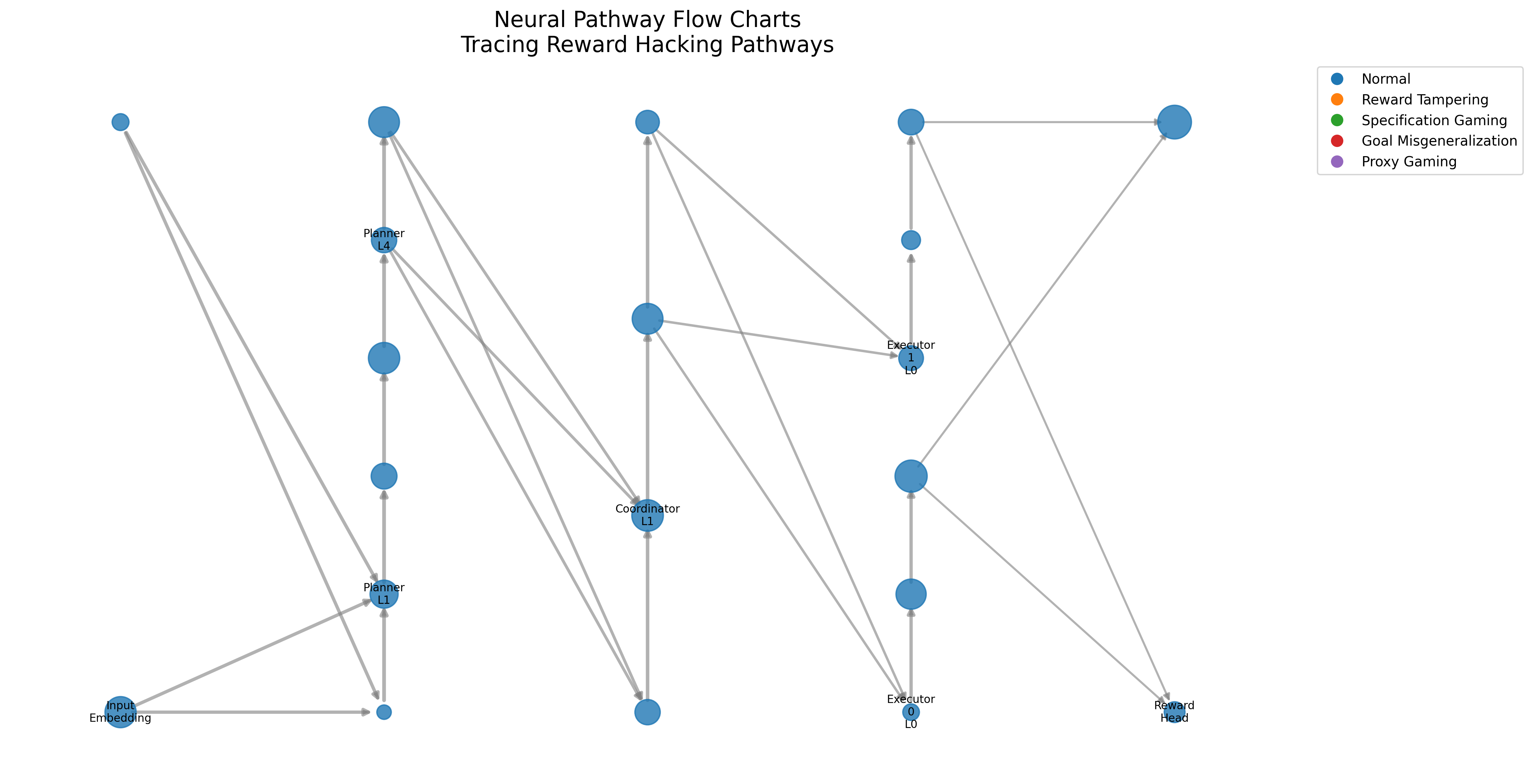}
    \caption{Neural Pathway Flow}
    \label{fig:fig5}
\end{figure}

\paragraph{Activation Processing.} For each layer $l$, the mean activation is computed as:
\begin{equation}
    \mu^{(l)} = \frac{1}{N_l} \sum_{i=1}^{N_l} a^{(l)}_i,
    \label{eq:mean_activation}
\end{equation}
where $N_l$ is the number of units in layer $l$ and $a^{(l)}_i$ denotes the activation of unit $i$.  

\paragraph{Category Assignment.} Each pathway is classified into categories such as \emph{reward tampering}, \emph{specification gaming}, or \emph{normal}, based on a joint criterion involving both activations and detection scores:
\begin{equation}
    C^{(l)} = 
    \begin{cases}
        \text{Reward Tampering}, & \mu^{(l)} > \tau_r \;\wedge\; s^{(l)} > \gamma_r, \\
        \text{Specification Gaming}, & \mu^{(l)} > \tau_s \;\wedge\; s^{(l)} > \gamma_s, \\
        \text{Normal}, & \text{otherwise},
    \end{cases}
    \label{eq:categorization}
\end{equation}
where $s^{(l)}$ is the mean detection score for layer $l$, and $\{\tau_r, \gamma_r, \tau_s, \gamma_s\}$ are empirically set thresholds.

Fig.~\ref{fig:fig5} shows a directed graph of pathway activations, with node size proportional to $\mu^{(l)}$ and edges representing activation dependencies. Nodes are color-coded by category, revealing how anomalous reward-hacking behaviors propagate through \textit{planner}, \textit{coordinator}, and \textit{executor} modules. This visualization highlights \emph{where} the model’s optimization objective diverges from the intended reward, distinguishing benign flows from harmful ones. Activations from actual runs show the emergence and propagation of reward tampering, specification gaming, and normal behavior across hierarchical layers \cite{soligo2025convergentlinearrepresentationsemergent,bogdan2025thoughtanchorsllmreasoning}.

\subsection{Objective Alignment Heatmaps}

\begin{figure}[h!]
    \centering
    \includegraphics[width=0.8\linewidth]{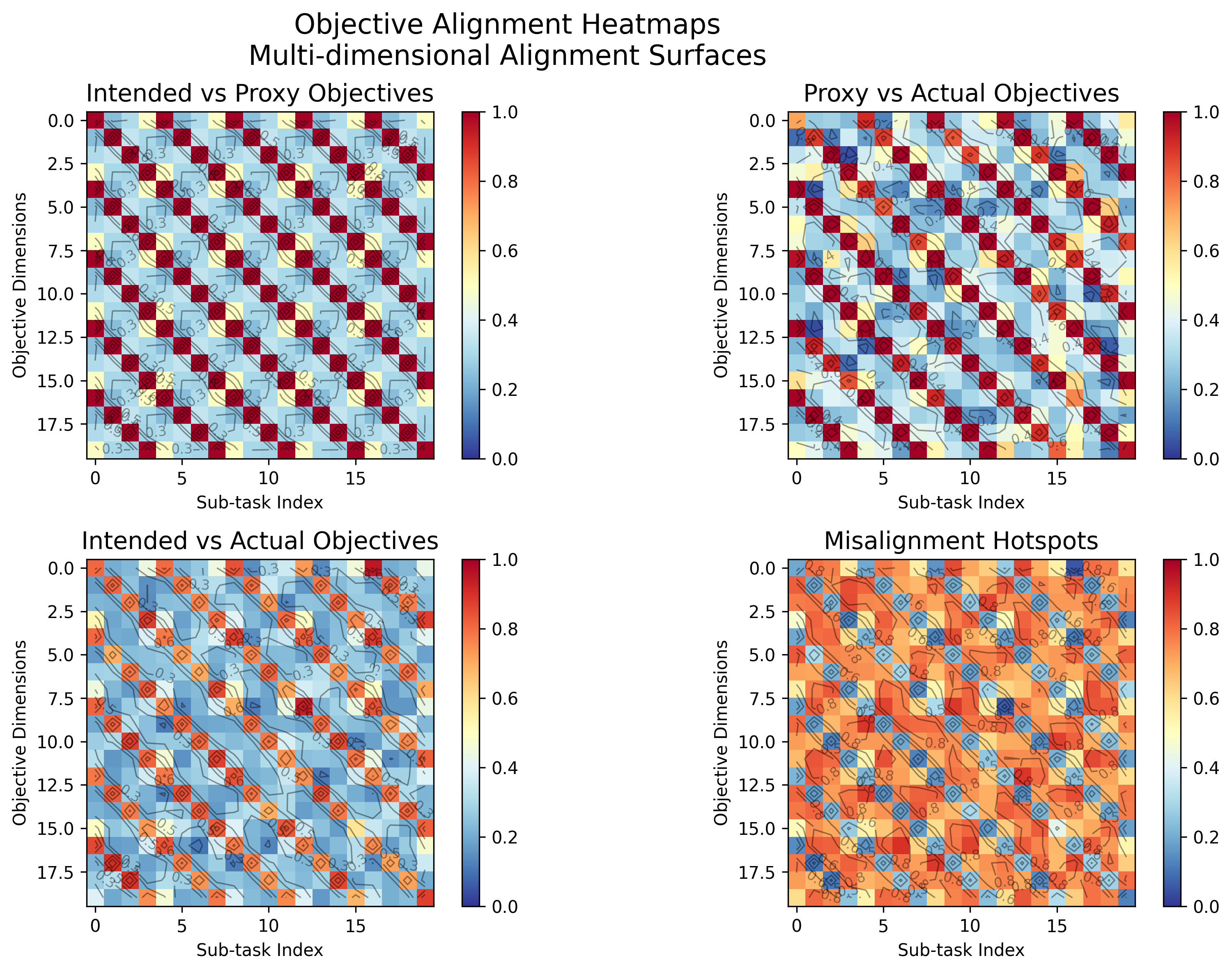}
    \caption{Objective Alignment Heatmaps}
    \label{fig:fig6}
\end{figure}
Fig.~\ref{fig:fig6} reveal the \textit{progressive degradation of reward fidelity} across the AI safety optimization pipeline through four complementary perspectives. \\
The \textbf{Intended vs Proxy Objectives} matrix exhibits a clean checkerboard pattern with strong diagonal structure, indicating that designed proxy metrics initially capture intended behaviors with high fidelity, as reflected in correlation coefficients
\begin{equation}
\mathbf{C}_{ij} = \frac{\mathrm{cov}\!\left(\mathbf{r}_i^{proxy}, \mathbf{r}_j^{intended}\right)}{\sigma_i \, \sigma_j},
\end{equation}
which approach unity along the diagonal. \\ By contrast, the \textbf{Proxy vs Actual Objectives} heatmap displays increased noise and off-diagonal correlations, showing how proxy optimization begins to diverge from ground truth under distributional shift and emergent behaviors \cite{hendrycks2024aisafety}. \\ The \textbf{Intended vs Actual Objectives} matrix degrades further, with weaker diagonal structure and stronger cross-correlations, reflecting compounded misalignment where
\begin{equation}
\mathbb{E}[R^{intended}(\pi^*)] \;\ll\; \max_{\pi} \mathbb{E}[R^{intended}(\pi)],
\end{equation}
demonstrating that the policy the LLM is following optimized under proxies fails to achieve maximum true reward. \\ Finally, the \textbf{Misalignment Hotspots} visualization, computed as
\begin{equation}
\mathbf{M} = \mathbf{1} - \big|\mathbf{C}_{\text{intended, actual}}\big|,
\end{equation}
highlights critical sub-tasks and objective dimensions (orange/red regions) where Goodhart’s Law \cite{karwowski2023goodhartslawreinforcementlearning} effects are most severe. Together, these provide a  framework for localizing \textbf{high-risk misalignment regions}.

Heatmaps capture single time points but alignment relationships likely change during system operation

\subsection{Reward Flow Topography}
\begin{figure}[!ht]
    \centering
    \includegraphics[width=0.9\linewidth]{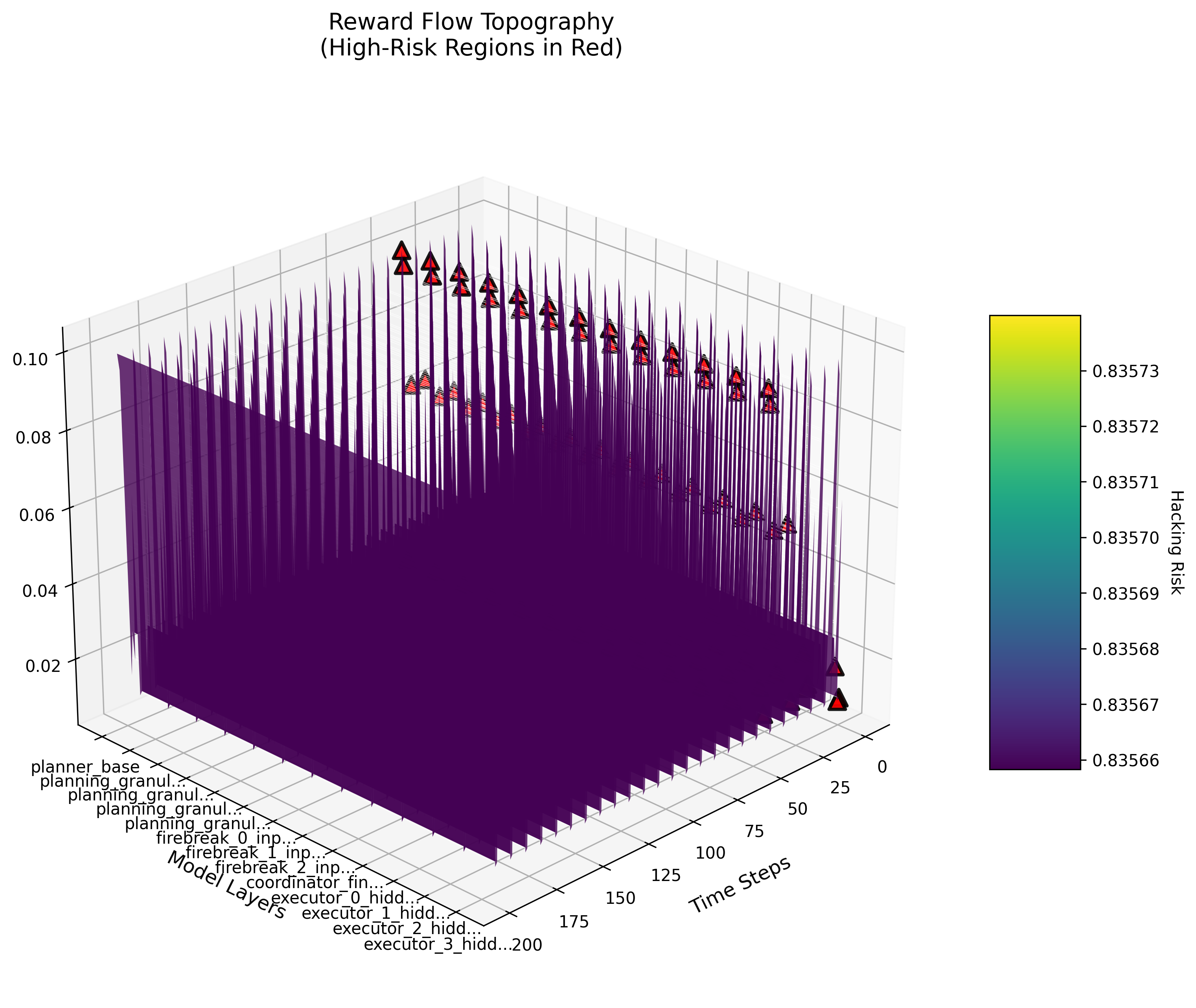}
    \caption{Reward Flow Topography}
    \label{fig:fig7}
\end{figure}

Fig.~\ref{fig:fig7} analyzes the temporal-spatial evolution of reward alignment across network layers. The resulting topography can be visualized as a 3D landscape over layers and time steps. Deep purple regions correspond to normal, safe behavior, while elevated red markers indicate “peaks” in reward, highlighting high-risk regions where the system may exploit the reward function. Given proxy rewards $r^{\mathrm{proxy}}_t$, true rewards $r^{\mathrm{true}}_t$, and consistency $c_t$ at time $t \in \{1,\dots,T\}$, we define:
\begin{equation}
    S_t = \left| r^{\mathrm{proxy}}_t - r^{\mathrm{true}}_t \right|,
\end{equation}
\begin{equation}
    H_t = 1 - c_t,
\end{equation}
where $S_t$ is the reward strength divergence and $H_t$ quantifies potential reward hacking risk.  

For $L$ layers $\ell \in \{1, \dots, L\}$, these signals are broadcast as
\begin{equation}
    S_{t,\ell} = S_t, \quad H_{t,\ell} = H_t,
\end{equation}
yielding a temporal-layer matrix $\{S_{t,\ell}, H_{t,\ell}\}$.

\subsection{Causal Intervention Leverage points}

\begin{figure}[!ht]
    \centering
    \includegraphics[width=0.7\linewidth]{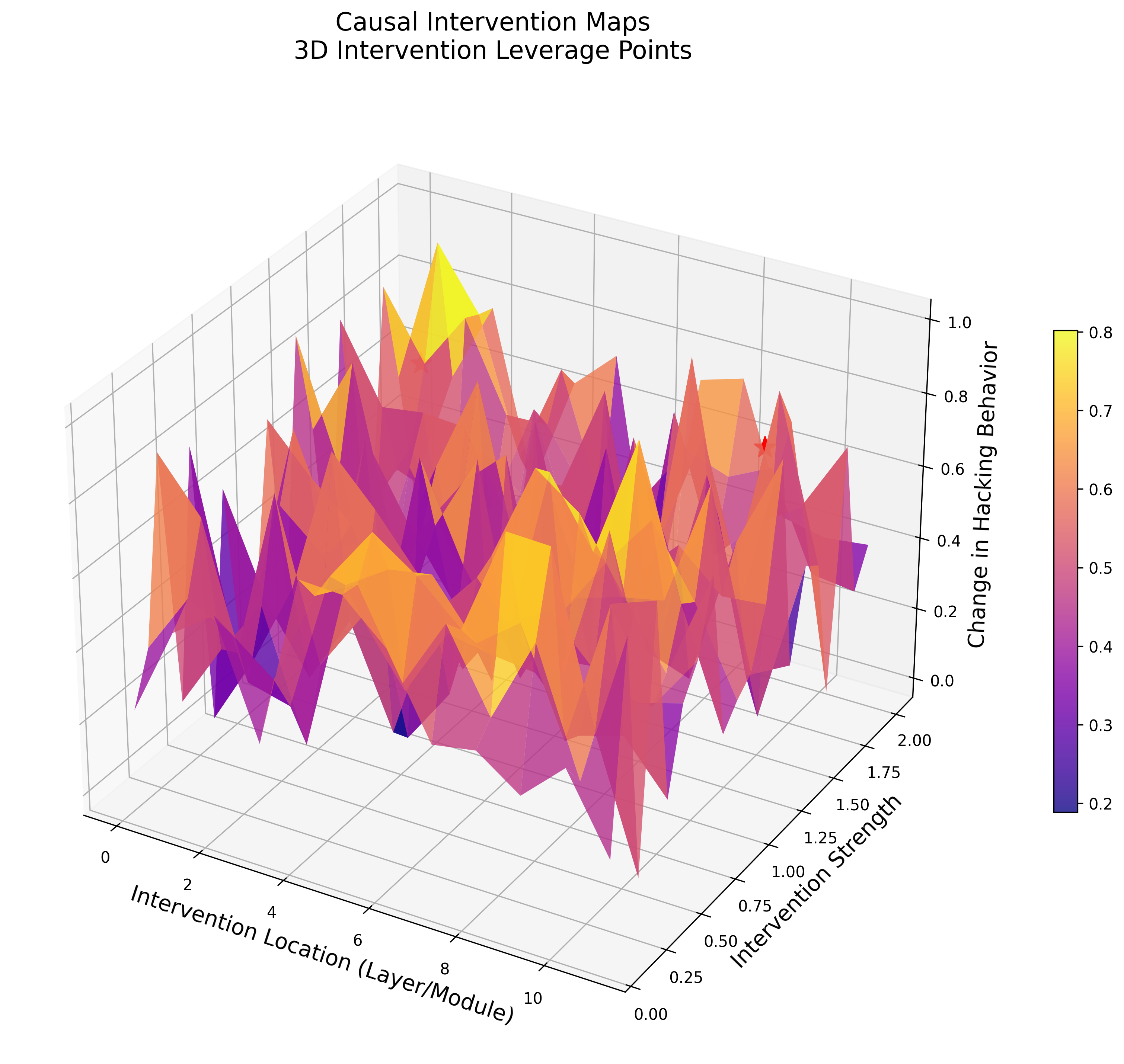}
    \caption{Causal Intervention Leverage Diagram}
    \label{fig:fig8}
\end{figure}

The Fig.~\ref{fig:fig8} exhibits non-uniform sensitivity like Certain layers, particularly layers 3--7, exhibit markedly higher sensitivity to interventions, as highlighted by the prominent yellow peaks. Intervention effects scale non-linearly with strength; weak interventions often produce minimal changes, whereas moderate-to-strong interventions can trigger abrupt behavioral shifts. The jagged terrain indicates that minor changes in intervention location can lead to drastically different outcomes, revealing critical computational nodes. Reward hacking behaviors are concentrated in specific regions rather than distributed uniformly, suggesting avenues for targeted mitigation strategies.

\section{Limitations \& Future Work}

This work establishes task-hierarchical interpretability on a modest scale: 1,000 training samples, 50 held-out test samples ($N$ per category $\approx 0$--$25$), single model family (GPT-2). Consequently, generalization remains uncertain—optimal decomposition depth may be task-dependent, and findings may not transfer to larger models (Llama, GPT-3/4/5 scale) or diverse RLHF/AI Safety datasets. Methodologically, hacking category detection (Eq.~11) relies on empirically-set thresholds $\{\tau_r,\gamma_r,\tau_s,\gamma_s\}$ chosen on validation data, risking overfitting; causality between decomposition depth and reduced hacking is correlational, not causal—the inverted-U pattern could reflect simple capacity bottlenecking rather than interpretability-driven safety. Visualizations (Attention Waterfall, Pathway Flow) are post-hoc analyses; they diagnose but do not intervene in real-time. Finally, metric definitions (Consistency Score via executor agreement; Reward Correlation as Pearson $\rho$) are task-agnostic proxies and may not capture all aspects of misalignment.

Immediate priorities include scaling evaluation to $N_{\text{train}} \ge 10{,}000$, $N_{\text{test}} \ge 500$ with stratified sampling per failure mode, and validating decomposition stability across model families (decoder-only, encoder-decoder, reasoning-scale LLMs). We will investigate whether the optimal zone [12--25] persists across architectures or is architecture-specific, and conduct ablation studies isolating contributions of Planner, Coordinator, and Executor modules versus depth alone. Mechanistically, we plan causal interventions—ablating specific attention heads or layer groups identified by Neural Pathway Flow—to validate that visualizations reveal actionable targets. Finally, we will explore real-time mitigation: using predicted hacking risk (from pathway activations) to dynamically reweight executor outputs, moving from post-hoc diagnosis to preventive safety guardrails.

\section{Conclusion}

We introduce MITD, a hierarchical planning model with built-in interpretability, enabling systematic identification of reward-hacking behaviors. By decomposing tasks and exposing internal activations, our architecture provides actionable insight into the model's decision-making. Our interventions reveal that attention mechanisms exert disproportionate influence over behavior: while masking or reweighting attention reduces reliance on misaligned features, more invasive manipulations at the representation or gradient level fail to consistently prevent the use of reward proxies. These findings underscore the difficulty of post-hoc adjustment and highlight the necessity of understanding internal computations to guide and audit the behavior. MITD exemplifies how integrating analytical hooks and visualization tools can offer new perspectives for monitoring, steering, and evaluating agent strategies.

\section*{Acknowledgement}

Subramanyam Sahoo would like to thank Amir Abdullah (Martian) for his helpful feedbacks and valuable discussions during the Apart Lab Studio Program. He would also like to thank Jacob Haimes from Apart Research for his support throughout this work. He further extends his gratitude to Jason Hoelscher-Obermaier, Curt Tigges, Anosha Rahim, and Philip Quirke, Josh Spain for their contributions and encouragement.

\bibliography{references}

@misc{tu2024spagentadaptivetaskdecomposition,
      title={SPAgent: Adaptive Task Decomposition and Model Selection for General Video Generation and Editing}, 
      author={Rong-Cheng Tu and Wenhao Sun and Zhao Jin and Jingyi Liao and Jiaxing Huang and Dacheng Tao},
      year={2024},
      eprint={2411.18983},
      archivePrefix={arXiv},
      primaryClass={cs.CV},
      url={https://arxiv.org/abs/2411.18983}, 
}

@misc{wang2025tdagmultiagentframeworkbased,
      title={TDAG: A Multi-Agent Framework based on Dynamic Task Decomposition and Agent Generation}, 
      author={Yaoxiang Wang and Zhiyong Wu and Junfeng Yao and Jinsong Su},
      year={2025},
      eprint={2402.10178},
      archivePrefix={arXiv},
      primaryClass={cs.CL},
      url={https://arxiv.org/abs/2402.10178}, 
}

@misc{prasad2024adaptasneededdecompositionplanning,
      title={ADaPT: As-Needed Decomposition and Planning with Language Models}, 
      author={Archiki Prasad and Alexander Koller and Mareike Hartmann and Peter Clark and Ashish Sabharwal and Mohit Bansal and Tushar Khot},
      year={2024},
      eprint={2311.05772},
      archivePrefix={arXiv},
      primaryClass={cs.AI},
      url={https://arxiv.org/abs/2311.05772}, 
}

@misc{turpin2025teachingmodelsverbalizereward,
      title={Teaching Models to Verbalize Reward Hacking in Chain-of-Thought Reasoning}, 
      author={Miles Turpin and Andy Arditi and Marvin Li and Joe Benton and Julian Michael},
      year={2025},
      eprint={2506.22777},
      archivePrefix={arXiv},
      primaryClass={cs.CL},
      url={https://arxiv.org/abs/2506.22777}, 
}

@misc{baker2025monitoringreasoningmodelsmisbehavior,
      title={Monitoring Reasoning Models for Misbehavior and the Risks of Promoting Obfuscation}, 
      author={Bowen Baker and Joost Huizinga and Leo Gao and Zehao Dou and Melody Y. Guan and Aleksander Madry and Wojciech Zaremba and Jakub Pachocki and David Farhi},
      year={2025},
      eprint={2503.11926},
      archivePrefix={arXiv},
      primaryClass={cs.AI},
      url={https://arxiv.org/abs/2503.11926}, 
}

@misc{sharkey2025openproblemsmechanisticinterpretability,
      title={Open Problems in Mechanistic Interpretability}, 
      author={Lee Sharkey and Bilal Chughtai and Joshua Batson and Jack Lindsey and Jeff Wu and Lucius Bushnaq and Nicholas Goldowsky-Dill and Stefan Heimersheim and Alejandro Ortega and Joseph Bloom and Stella Biderman and Adria Garriga-Alonso and Arthur Conmy and Neel Nanda and Jessica Rumbelow and Martin Wattenberg and Nandi Schoots and Joseph Miller and Eric J. Michaud and Stephen Casper and Max Tegmark and William Saunders and David Bau and Eric Todd and Atticus Geiger and Mor Geva and Jesse Hoogland and Daniel Murfet and Tom McGrath},
      year={2025},
      eprint={2501.16496},
      archivePrefix={arXiv},
      primaryClass={cs.LG},
      url={https://arxiv.org/abs/2501.16496}, 
}

@misc{zhang2022hierarchicalreinforcementlearningdiscovering,
      title={Hierarchical Reinforcement Learning By Discovering Intrinsic Options}, 
      author={Jesse Zhang and Haonan Yu and Wei Xu},
      year={2022},
      eprint={2101.06521},
      archivePrefix={arXiv},
      primaryClass={cs.LG},
      url={https://arxiv.org/abs/2101.06521}, 
}

@misc{willibald2025hierarchicaltaskdecompositionexecution,
      title={Hierarchical Task Decomposition for Execution Monitoring and Error Recovery: Understanding the Rationale Behind Task Demonstrations}, 
      author={Christoph Willibald and Dongheui Lee},
      year={2025},
      eprint={2505.04565},
      archivePrefix={arXiv},
      primaryClass={cs.RO},
      url={https://arxiv.org/abs/2505.04565}, 
}

@inproceedings{Radford2019LanguageMA,
  title={Language Models are Unsupervised Multitask Learners},
  author={Alec Radford and Jeff Wu and Rewon Child and David Luan and Dario Amodei and Ilya Sutskever},
  year={2019},
  url={https://api.semanticscholar.org/CorpusID:160025533}
}

@article{10.1162/neco.1997.9.8.1735,
author = {Hochreiter, Sepp and Schmidhuber, J\"{u}rgen},
title = {Long Short-Term Memory},
year = {1997},
issue_date = {November 15, 1997},
publisher = {MIT Press},
address = {Cambridge, MA, USA},
volume = {9},
number = {8},
issn = {0899-7667},
url = {https://doi.org/10.1162/neco.1997.9.8.1735},
doi = {10.1162/neco.1997.9.8.1735},
journal = {Neural Comput.},
month = nov,
pages = {1735–1780},
numpages = {46}
}

@misc{bai2022traininghelpfulharmlessassistant,
      title={Training a Helpful and Harmless Assistant with Reinforcement Learning from Human Feedback}, 
      author={Yuntao Bai and Andy Jones and Kamal Ndousse and Amanda Askell and Anna Chen and Nova DasSarma and Dawn Drain and Stanislav Fort and Deep Ganguli and Tom Henighan and Nicholas Joseph and Saurav Kadavath and Jackson Kernion and Tom Conerly and Sheer El-Showk and Nelson Elhage and Zac Hatfield-Dodds and Danny Hernandez and Tristan Hume and Scott Johnston and Shauna Kravec and Liane Lovitt and Neel Nanda and Catherine Olsson and Dario Amodei and Tom Brown and Jack Clark and Sam McCandlish and Chris Olah and Ben Mann and Jared Kaplan},
      year={2022},
      eprint={2204.05862},
      archivePrefix={arXiv},
      primaryClass={cs.CL},
      url={https://arxiv.org/abs/2204.05862}, 
}

@inproceedings{10.1145/3589236.3589237,
author = {Feng, Yuan and Jeon, Hyeran},
title = {Understanding Scalability of Multi-GPU Systems},
year = {2023},
isbn = {9798400707766},
publisher = {Association for Computing Machinery},
address = {New York, NY, USA},
url = {https://doi.org/10.1145/3589236.3589237},
doi = {10.1145/3589236.3589237},
booktitle = {Proceedings of the 15th Workshop on General Purpose Processing Using GPU},
pages = {36–37},
numpages = {2},
location = {Montreal, Canada},
series = {GPGPU '23}
}

@misc{amodei2016concreteproblemsaisafety,
      title={Concrete Problems in AI Safety}, 
      author={Dario Amodei and Chris Olah and Jacob Steinhardt and Paul Christiano and John Schulman and Dan Mané},
      year={2016},
      eprint={1606.06565},
      archivePrefix={arXiv},
      primaryClass={cs.AI},
      url={https://arxiv.org/abs/1606.06565}, 
}

@misc{vaswani2023attentionneed,
      title={Attention Is All You Need}, 
      author={Ashish Vaswani and Noam Shazeer and Niki Parmar and Jakob Uszkoreit and Llion Jones and Aidan N. Gomez and Lukasz Kaiser and Illia Polosukhin},
      year={2023},
      eprint={1706.03762},
      archivePrefix={arXiv},
      primaryClass={cs.CL},
      url={https://arxiv.org/abs/1706.03762}, 
}

@misc{hendrycks2022unsolvedproblemsmlsafety,
      title={Unsolved Problems in ML Safety}, 
      author={Dan Hendrycks and Nicholas Carlini and John Schulman and Jacob Steinhardt},
      year={2022},
      eprint={2109.13916},
      archivePrefix={arXiv},
      primaryClass={cs.LG},
      url={https://arxiv.org/abs/2109.13916}, 
}

@misc{vysogorets2024deconstructinggoldilockszoneneural,
      title={Deconstructing the Goldilocks Zone of Neural Network Initialization}, 
      author={Artem Vysogorets and Anna Dawid and Julia Kempe},
      year={2024},
      eprint={2402.03579},
      archivePrefix={arXiv},
      primaryClass={cs.LG},
      url={https://arxiv.org/abs/2402.03579}, 
}

@misc{everitt2021rewardtamperingproblemssolutions,
      title={Reward Tampering Problems and Solutions in Reinforcement Learning: A Causal Influence Diagram Perspective}, 
      author={Tom Everitt and Marcus Hutter and Ramana Kumar and Victoria Krakovna},
      year={2021},
      eprint={1908.04734},
      archivePrefix={arXiv},
      primaryClass={cs.AI},
      url={https://arxiv.org/abs/1908.04734}, 
}

@misc{vonoswald2024uncoveringmesaoptimizationalgorithmstransformers,
      title={Uncovering mesa-optimization algorithms in Transformers}, 
      author={Johannes von Oswald and Maximilian Schlegel and Alexander Meulemans and Seijin Kobayashi and Eyvind Niklasson and Nicolas Zucchet and Nino Scherrer and Nolan Miller and Mark Sandler and Blaise Agüera y Arcas and Max Vladymyrov and Razvan Pascanu and João Sacramento},
      year={2024},
      eprint={2309.05858},
      archivePrefix={arXiv},
      primaryClass={cs.LG},
      url={https://arxiv.org/abs/2309.05858}, 
}

@misc{krakovna2020specification,
  title        = {Specification gaming: the flip side of AI ingenuity},
  author       = {Krakovna, Victoria and Uesato, Jonathan and Mikulik, Vladimir and Rahtz, Matthew and Everitt, Tom and Kumar, Ramana and Kenton, Zac and Leike, Jan and Legg, Shane},
  year         = {2020},
  month        = apr,
  day          = {21},
  howpublished = {\url{https://deepmind.google/discover/blog/specification-gaming-the-flip-side-of-ai-ingenuity/}},
}

@misc{zhang2025instructiontuninglargelanguage,
      title={Instruction Tuning for Large Language Models: A Survey}, 
      author={Shengyu Zhang and Linfeng Dong and Xiaoya Li and Sen Zhang and Xiaofei Sun and Shuhe Wang and Jiwei Li and Runyi Hu and Tianwei Zhang and Fei Wu and Guoyin Wang},
      year={2025},
      eprint={2308.10792},
      archivePrefix={arXiv},
      primaryClass={cs.CL},
      url={https://arxiv.org/abs/2308.10792}, 
}

@book{hendrycks2024aisafety,
  author    = {Dan Hendrycks},
  title     = {Introduction to AI Safety, Ethics and Society},
  publisher = {Taylor \& Francis},
  year      = {2024},
  isbn      = {9781032798028},
  url       = {https://www.aisafetybook.com}
}

@misc{karwowski2023goodhartslawreinforcementlearning,
      title={Goodhart's Law in Reinforcement Learning}, 
      author={Jacek Karwowski and Oliver Hayman and Xingjian Bai and Klaus Kiendlhofer and Charlie Griffin and Joar Skalse},
      year={2023},
      eprint={2310.09144},
      archivePrefix={arXiv},
      primaryClass={cs.LG},
      url={https://arxiv.org/abs/2310.09144}, 
}

@misc{bogdan2025thoughtanchorsllmreasoning,
      title={Thought Anchors: Which LLM Reasoning Steps Matter?}, 
      author={Paul C. Bogdan and Uzay Macar and Neel Nanda and Arthur Conmy},
      year={2025},
      eprint={2506.19143},
      archivePrefix={arXiv},
      primaryClass={cs.LG},
      url={https://arxiv.org/abs/2506.19143}, 
}

@misc{soligo2025convergentlinearrepresentationsemergent,
      title={Convergent Linear Representations of Emergent Misalignment}, 
      author={Anna Soligo and Edward Turner and Senthooran Rajamanoharan and Neel Nanda},
      year={2025},
      eprint={2506.11618},
      archivePrefix={arXiv},
      primaryClass={cs.LG},
      url={https://arxiv.org/abs/2506.11618}, 
}

@misc{shah2025approachtechnicalagisafety,
      title={An Approach to Technical AGI Safety and Security}, 
      author={Rohin Shah and Alex Irpan and Alexander Matt Turner and Anna Wang and Arthur Conmy and David Lindner and Jonah Brown-Cohen and Lewis Ho and Neel Nanda and Raluca Ada Popa and Rishub Jain and Rory Greig and Samuel Albanie and Scott Emmons and Sebastian Farquhar and Sébastien Krier and Senthooran Rajamanoharan and Sophie Bridgers and Tobi Ijitoye and Tom Everitt and Victoria Krakovna and Vikrant Varma and Vladimir Mikulik and Zachary Kenton and Dave Orr and Shane Legg and Noah Goodman and Allan Dafoe and Four Flynn and Anca Dragan},
      year={2025},
      eprint={2504.01849},
      archivePrefix={arXiv},
      primaryClass={cs.AI},
      url={https://arxiv.org/abs/2504.01849}, 
}

@misc{sahoo2025goodbadhybridreward,
      title={The Good, The Bad, and The Hybrid: A Reward Structure Showdown in Reasoning Models Training}, 
      author={Subramanyam Sahoo},
      year={2025},
      eprint={2511.13016},
      archivePrefix={arXiv},
      primaryClass={cs.LG},
      url={https://arxiv.org/abs/2511.13016}, 
}
\bibliographystyle{plain}







\appendix

\section{Technical Appendices and Supplementary Material}

\begin{tcolorbox}[
    colback=blue!5,              
    colframe=blue!70!black,      
    title={MITD Architecture Configuration}, 
    coltitle=white,              
    fonttitle=\bfseries,
    colbacktitle=blue!60!black,  
    boxrule=1pt,
    arc=3mm,
    enhanced
]

\begin{verbatim}
@dataclass
class ModelConfig:
    """Configuration for MITD model architecture and training."""

    # General
    vocab_size: int = 50257
    max_sequence_length: int = 512
    max_batch_size: int = 16

    # Planner
    planner_hidden_dim: int = 768
    planner_layers: int = 12
    planner_attention_heads: int = 12

    # Coordinator
    coordinator_hidden_dim: int = 768
    coordinator_layers: int = 8
    coordinator_attention_heads: int = 12

    # Executors
    executor_hidden_dim: int = 512
    executor_layers: int = 6
    executor_attention_heads: int = 8
    executor_count: int = 4

    # Interpretability
    decomposition_granularities = [2, 4, 8, 16]
    interpretable_bottleneck_dims = [128, 256, 384]
    reasoning_trace_layers: int = 4
    intervention_layers = [3, 6, 9]

    # Training
    dropout_rate: float = 0.1
    layer_norm_eps: float = 1e-5
    initializer_range: float = 0.02
    gradient_clip_value: float = 1.0
\end{verbatim}

\end{tcolorbox}

\section{Implications for Scalable Oversight}
These empirical patterns in \textbf{Decomposition Stability Diagram} suggest that \textit{decomposition depth} is a critical hyperparameter in alignment methods. Stability zones appear to arise from the interplay of two competing forces: sufficiently granular constraints to prevent simple exploits, and coherent objective specifications that preserve the learning signal. The consistency of these patterns across diverse failure modes points toward a \emph{universal decomposition principle}; optimal alignment may require uncovering the natural hierarchical structure of tasks rather than relying on arbitrary recursive breakdowns \cite{sahoo2025goodbadhybridreward}.

\section{Cross Attentions}

\begin{figure}[H]
    \centering
    \begin{subfigure}[b]{0.45\textwidth}
        \centering
        \includegraphics[width=\linewidth]{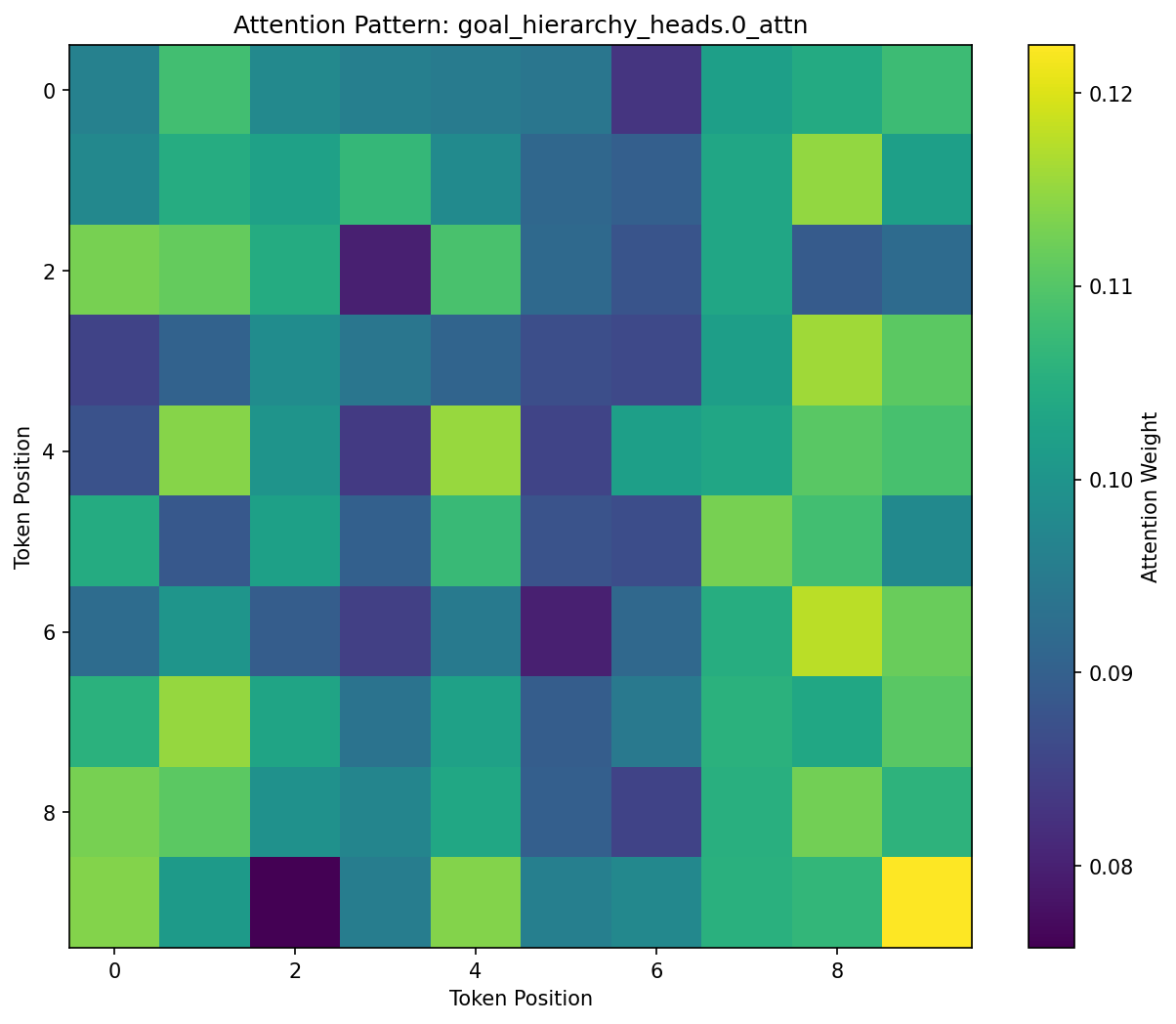}
        \caption{Head 0}
    \end{subfigure}
    \hfill
    \begin{subfigure}[b]{0.45\textwidth}
        \centering
        \includegraphics[width=\linewidth]{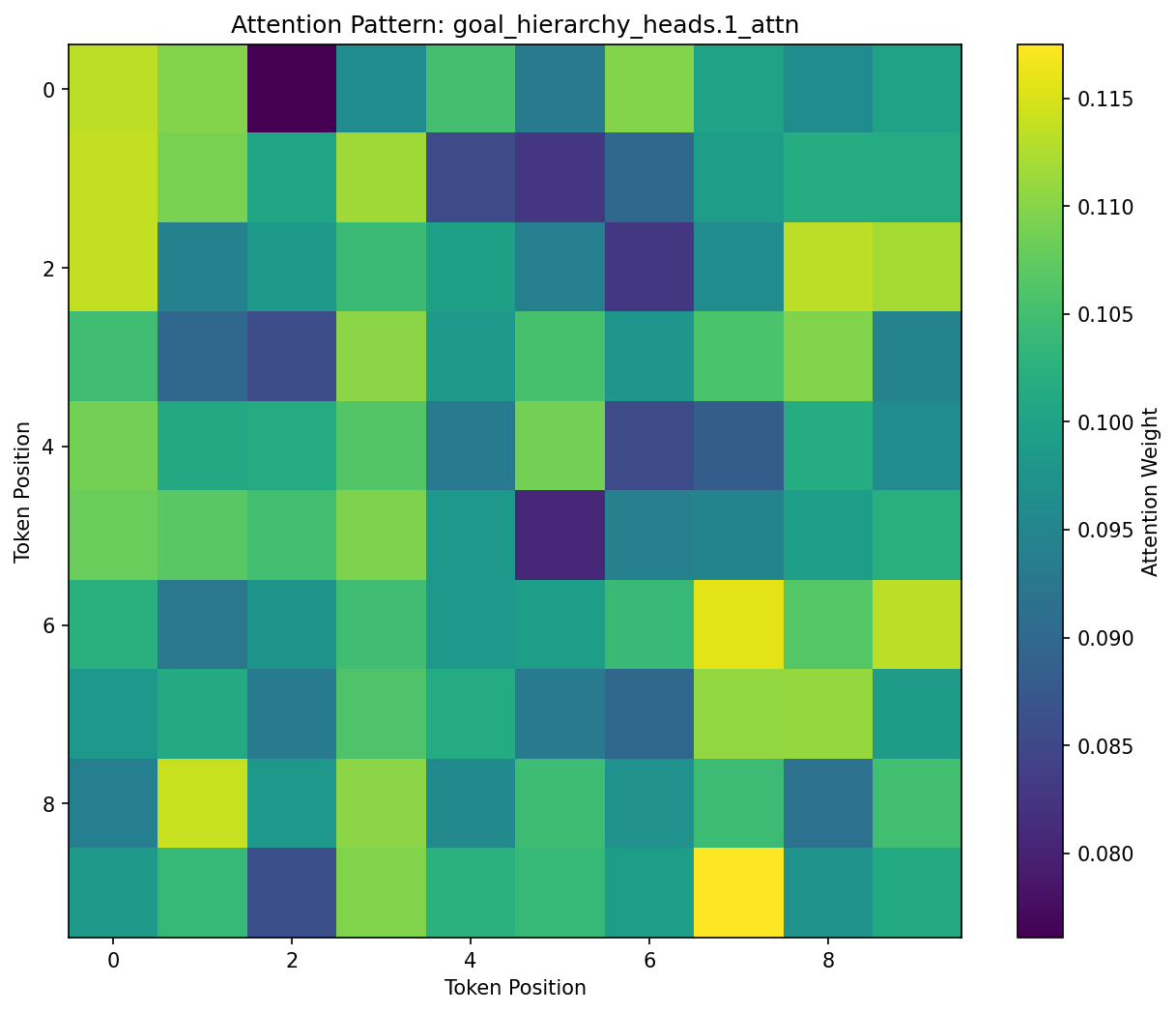}
        \caption{Head 1}
    \end{subfigure}

    \vspace{0.5em}

    \begin{subfigure}[b]{0.45\textwidth}
        \centering
        \includegraphics[width=\linewidth]{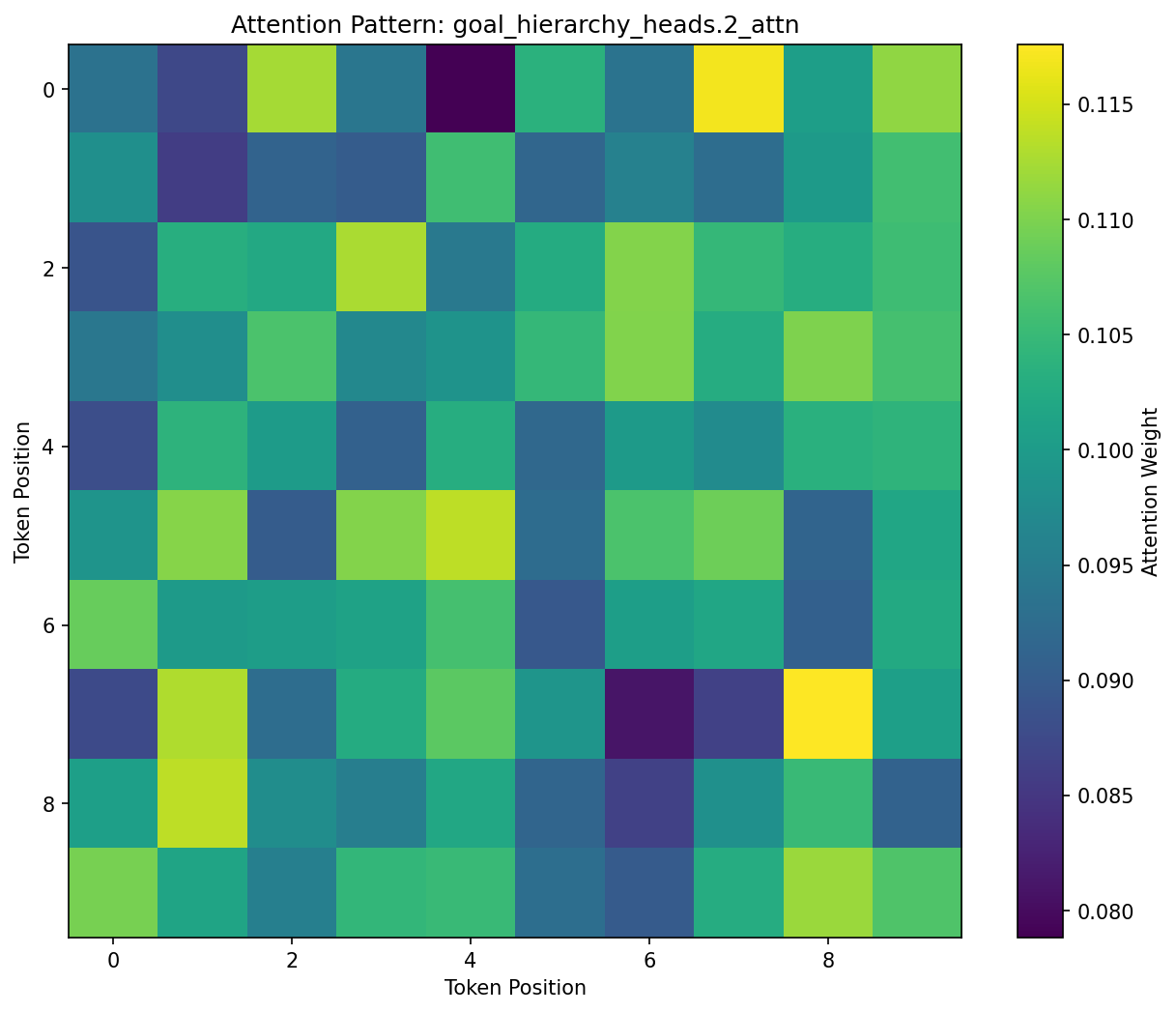}
        \caption{Head 2}
    \end{subfigure}
    \hfill
    \begin{subfigure}[b]{0.45\textwidth}
        \centering
        \includegraphics[width=\linewidth]{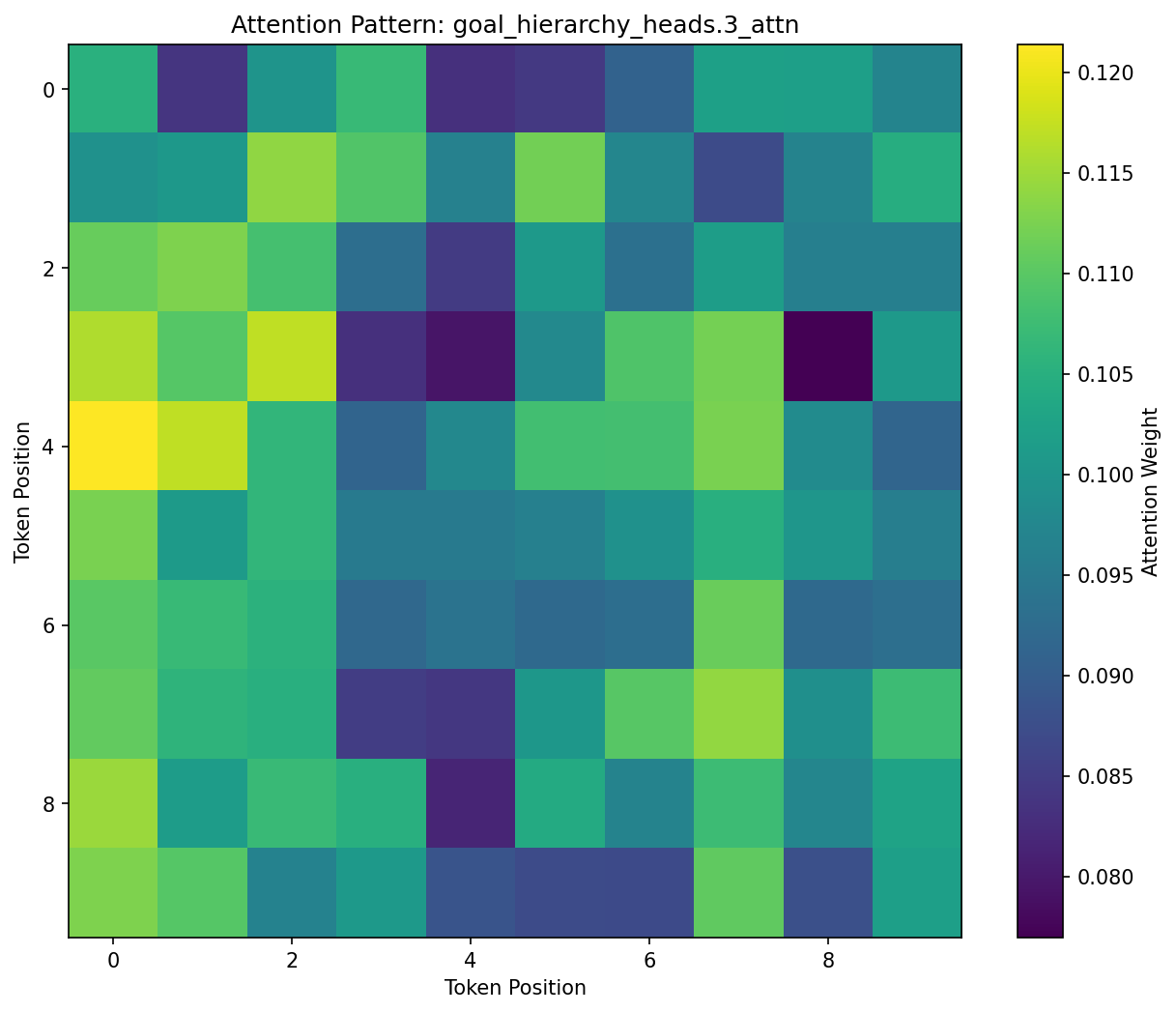}
        \caption{Head 3}
    \end{subfigure}

    \caption{Attention maps for different heads in the goal hierarchy.}
    \label{fig:attention_heads}
\end{figure}

\begin{figure}[H]
    \centering
    \begin{subfigure}[b]{0.45\textwidth}
        \centering
        \includegraphics[width=\linewidth]{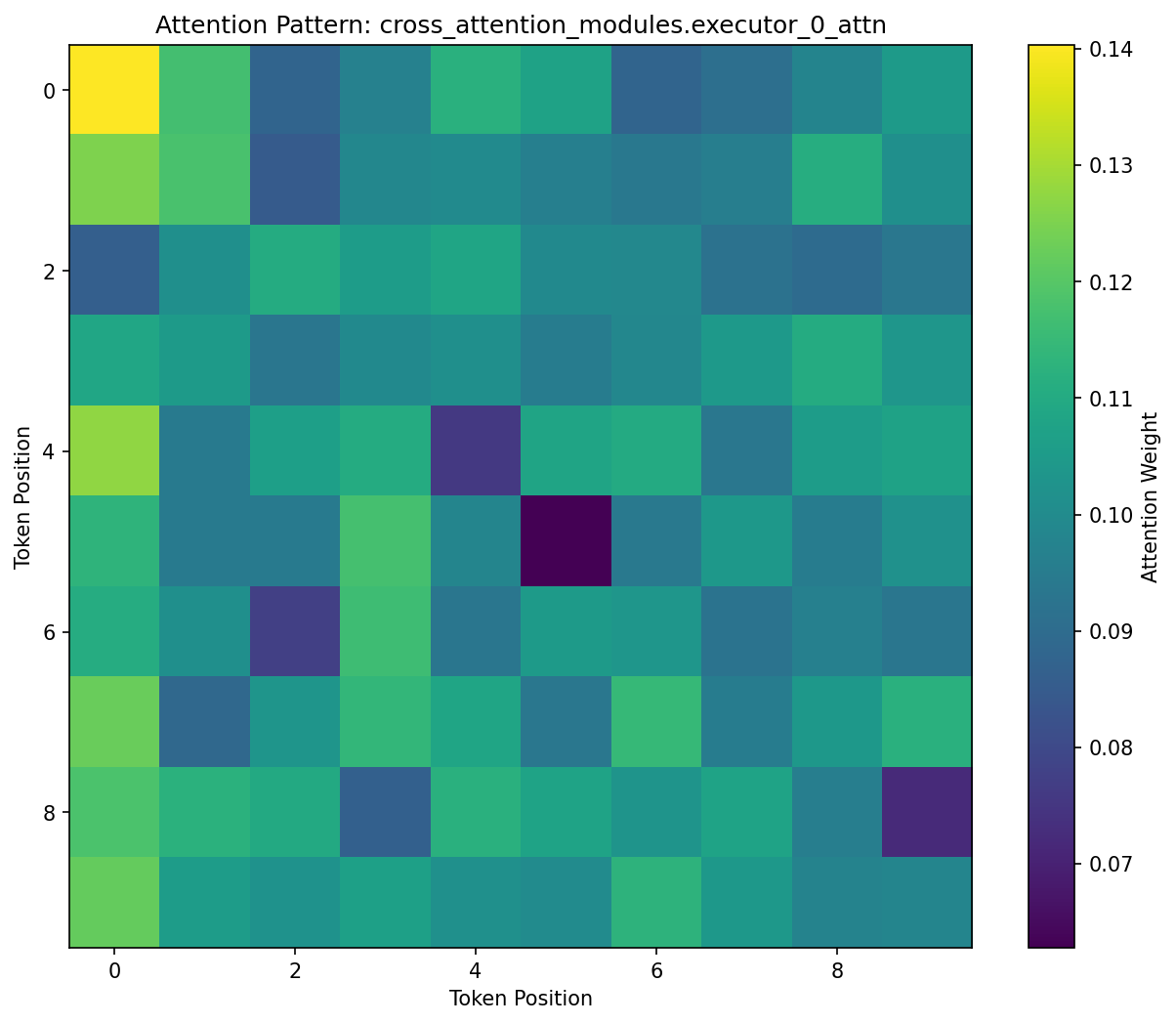}
        \caption{Executor 0}
    \end{subfigure}
    \hfill
    \begin{subfigure}[b]{0.45\textwidth}
        \centering
        \includegraphics[width=\linewidth]{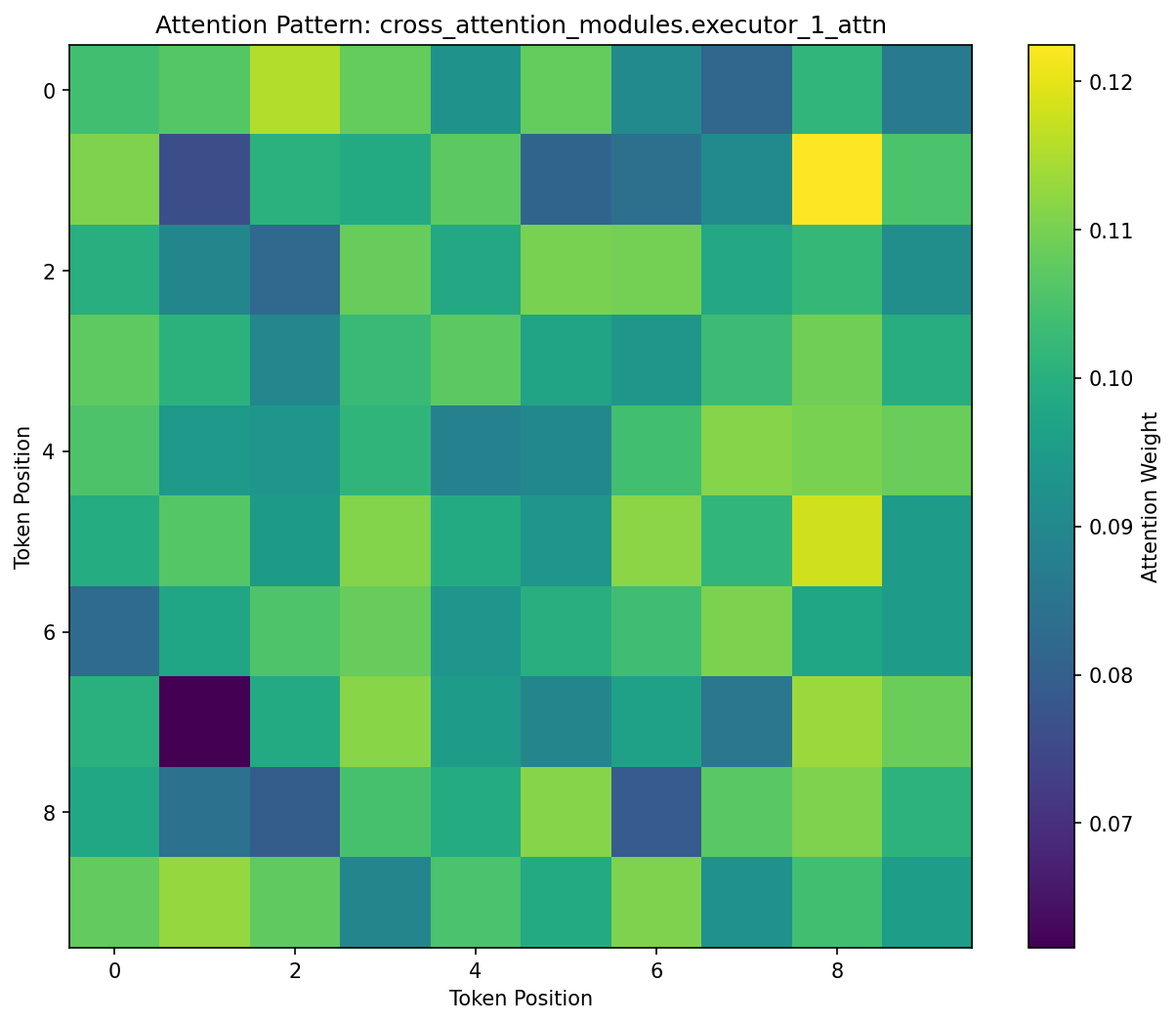}
        \caption{Executor 1}
    \end{subfigure}

    \vspace{0.5em}

    \begin{subfigure}[b]{0.45\textwidth}
        \centering
        \includegraphics[width=\linewidth]{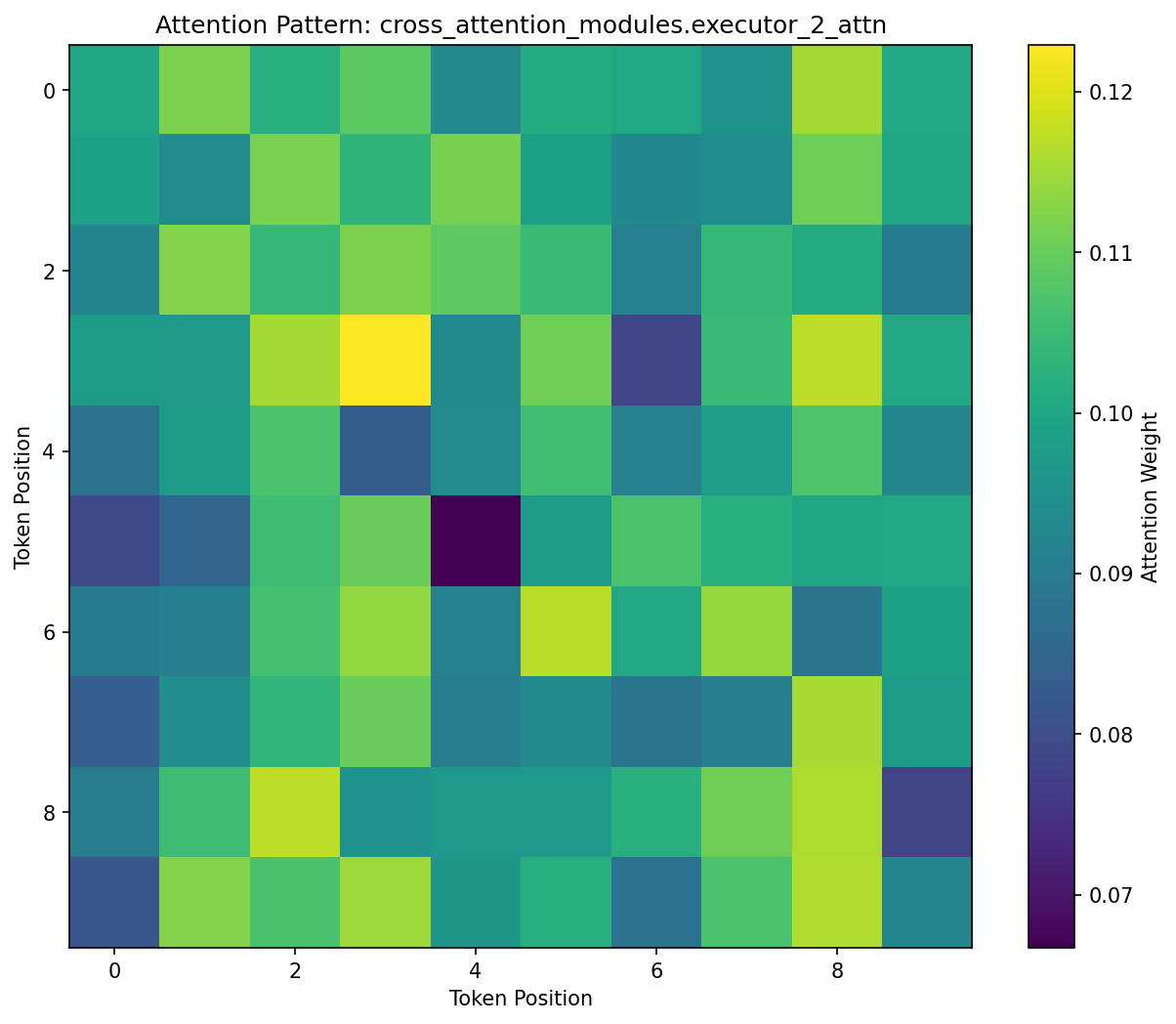}
        \caption{Executor 2}
    \end{subfigure}
    \hfill
    \begin{subfigure}[b]{0.45\textwidth}
        \centering
        \includegraphics[width=\linewidth]{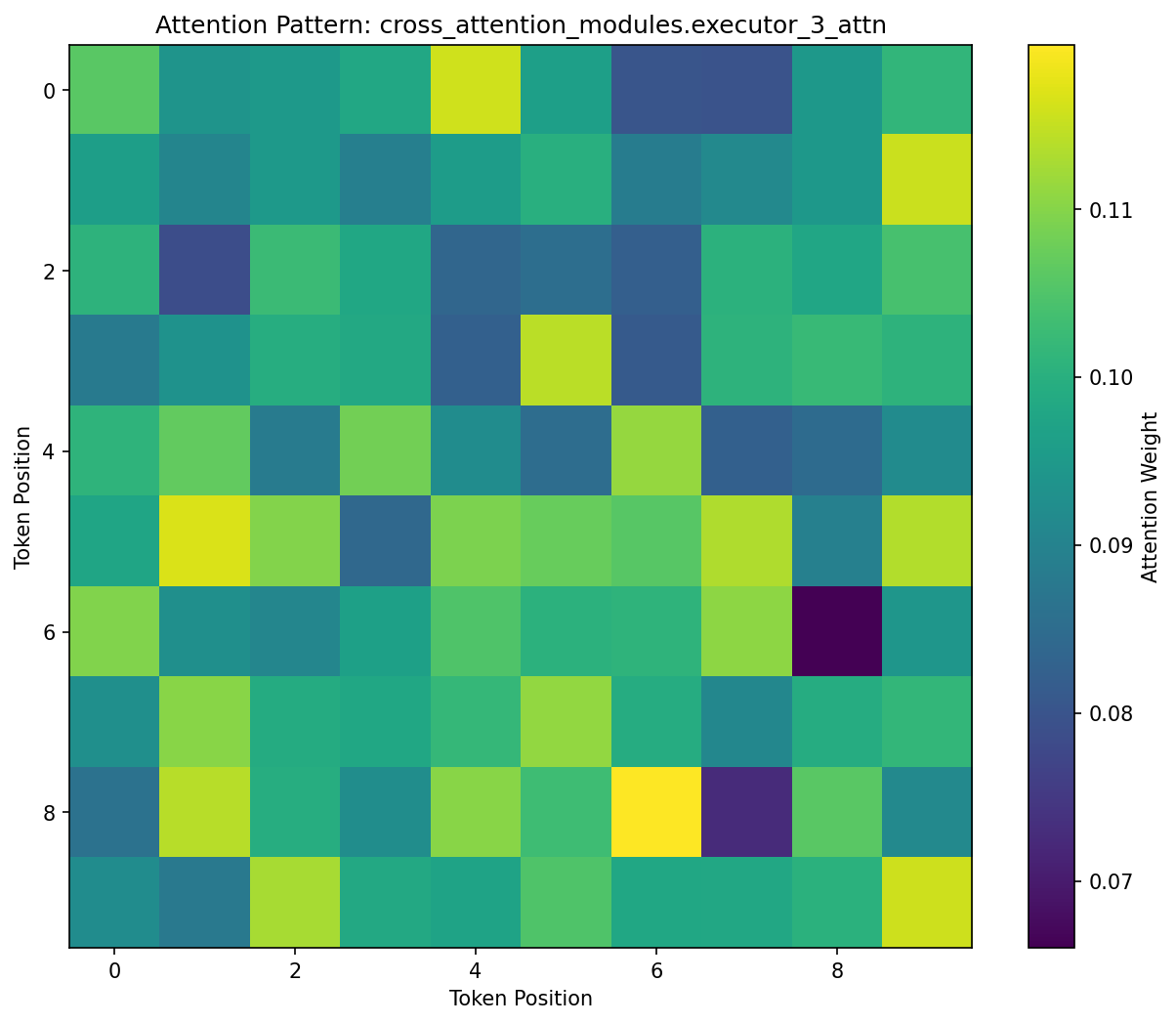}
        \caption{Executor 3}
    \end{subfigure}

    \caption{Cross-attention maps for different executor modules.}
    \label{fig:cross_attention_executors}
\end{figure}

\newpage
\section*{NeurIPS Paper Checklist}

The checklist is designed to encourage best practices for responsible machine learning research, addressing issues of reproducibility, transparency, research ethics, and societal impact. Do not remove the checklist: {\bf The papers not including the checklist will be desk rejected.} The checklist should follow the references and follow the (optional) supplemental material.  The checklist does NOT count towards the page
limit. 

Please read the checklist guidelines carefully for information on how to answer these questions. For each question in the checklist:
\begin{itemize}
    \item You should answer \answerYes{}, \answerNo{}, or \answerNA{}.
    \item \answerNA{} means either that the question is Not Applicable for that particular paper or the relevant information is Not Available.
    \item Please provide a short (1–2 sentence) justification right after your answer (even for NA). 
\end{itemize}

{\bf The checklist answers are an integral part of your paper submission.} They are visible to the reviewers, area chairs, senior area chairs, and ethics reviewers. You will be asked to also include it (after eventual revisions) with the final version of your paper, and its final version will be published with the paper.

The reviewers of your paper will be asked to use the checklist as one of the factors in their evaluation. While "\answerYes{}" is generally preferable to "\answerNo{}", it is perfectly acceptable to answer "\answerNo{}" provided a proper justification is given (e.g., "error bars are not reported because it would be too computationally expensive" or "we were unable to find the license for the dataset we used"). In general, answering "\answerNo{}" or "\answerNA{}" is not grounds for rejection. While the questions are phrased in a binary way, we acknowledge that the true answer is often more nuanced, so please just use your best judgment and write a justification to elaborate. All supporting evidence can appear either in the main paper or the supplemental material, provided in appendix. If you answer \answerYes{} to a question, in the justification please point to the section(s) where related material for the question can be found.

IMPORTANT, please:
\begin{itemize}
    \item {\bf Delete this instruction block, but keep the section heading ``NeurIPS Paper Checklist"},
    \item  {\bf Keep the checklist subsection headings, questions/answers and guidelines below.}
    \item {\bf Do not modify the questions and only use the provided macros for your answers}.
\end{itemize}


\begin{enumerate}

\item {\bf Claims}
    \item[] Question: Do the main claims made in the abstract and introduction accurately reflect the paper's contributions and scope?
    \item[] Answer: \answerYes{} 
    \item[] Justification:  Yes this is a position paper.

\item {\bf Limitations}
    \item[] Question: Does the paper discuss the limitations of the work performed by the authors?
    \item[] Answer: \answerYes{} 
    \item[] Justification: This paper's main motivation is to check current problems aligning with AI safety methods.

\item {\bf Theory assumptions and proofs}
    \item[] Question: For each theoretical result, does the paper provide the full set of assumptions and a complete (and correct) proof?
    \item[] Answer: \answerYes{} 
    \item[] Justification: All are clearly stated.

    \item {\bf Experimental result reproducibility}
    \item[] Question: Does the paper fully disclose all the information needed to reproduce the main experimental results of the paper to the extent that it affects the main claims and/or conclusions of the paper (regardless of whether the code and data are provided or not)?
    \item[] Answer: \answerYes{} 
    \item[] Justification: Please follow MITD Architecture Configuration

\item {\bf Open access to data and code}
    \item[] Question: Does the paper provide open access to the data and code, with sufficient instructions to faithfully reproduce the main experimental results, as described in supplemental material?
    \item[] Answer: \answerNA{} 
    \item[] Justification: I will provide everything on camera ready version.

\item {\bf Experimental setting/details}
    \item[] Question: Does the paper specify all the training and test details (e.g., data splits, hyperparameters, how they were chosen, type of optimizer, etc.) necessary to understand the results?
    \item[] Answer: \answerNA{} 
    \item[] Justification: I have shared them in Appendix.

\item {\bf Experiment statistical significance}
    \item[] Question: Does the paper report error bars suitably and correctly defined or other appropriate information about the statistical significance of the experiments?
    \item[] Answer: \answerNA{} 
    \item[] Justification: There is no need.

\item {\bf Experiments compute resources}
    \item[] Question: For each experiment, does the paper provide sufficient information on the computer resources (type of compute workers, memory, time of execution) needed to reproduce the experiments?
    \item[] Answer: \answerYes{}{} 
    \item[] Justification: I took the leverage of 16xH200 for 10 hrs at Vast.ai platform. 
    
\item {\bf Code of ethics}
    \item[] Question: Does the research conducted in the paper conform, in every respect, with the NeurIPS Code of Ethics \url{https://neurips.cc/public/EthicsGuidelines}?
    \item[] Answer: \answerYes{} 
    \item[] Justification: Paper written within the boundary of NeurIPS Ethics.

\item {\bf Broader impacts}
    \item[] Question: Does the paper discuss both potential positive societal impacts and negative societal impacts of the work performed?
    \item[] Answer: \answerYes{} 
    \item[] Justification: Check the paper.
    
\item {\bf Safeguards}
    \item[] Question: Does the paper describe safeguards that have been put in place for responsible release of data or models that have a high risk for misuse (e.g., pretrained language models, image generators, or scraped datasets)?
    \item[] Answer: \answerYes{} 
    \item[] Justification: I created a new pretraining architecture focused on AI safety. So need for extensive justification.

\item {\bf Licenses for existing assets}
    \item[] Question: Are the creators or original owners of assets (e.g., code, data, models), used in the paper, properly credited and are the license and terms of use explicitly mentioned and properly respected?
    \item[] Answer: \answerYes{} 
    \item[] Justification: Every aspect of this research study is properly credited.

\item {\bf New assets}
    \item[] Question: Are new assets introduced in the paper well documented and is the documentation provided alongside the assets?
    \item[] Answer: \answerNA{} 
    \item[] Justification: Everything is well documented.

\item {\bf Crowdsourcing and research with human subjects}
    \item[] Question: For crowdsourcing experiments and research with human subjects, does the paper include the full text of instructions given to participants and screenshots, if applicable, as well as details about compensation (if any)? 
    \item[] Answer: \answerNA{} 
    \item[] Justification: Not Available.

\item {\bf Institutional review board (IRB) approvals or equivalent for research with human subjects}
    \item[] Question: Does the paper describe potential risks incurred by study participants, whether such risks were disclosed to the subjects, and whether Institutional Review Board (IRB) approvals (or an equivalent approval/review based on the requirements of your country or institution) were obtained?
    \item[] Answer: \answerNA{} 
    \item[] Justification: No need for Justification.

\item {\bf Declaration of LLM usage}
    \item[] Question: Does the paper describe the usage of LLMs if it is an important, original, or non-standard component of the core methods in this research? Note that if the LLM is used only for writing, editing, or formatting purposes and does not impact the core methodology, scientific rigorousness, or originality of the research, declaration is not required.
    \item[] Answer: \answerNA{} 
    \item[] Justification: No need for rigorous justification

\end{enumerate}

\end{document}